\pdfoutput=1
\def\year{2017}\relax
\documentclass[letterpaper]{article}
\usepackage{aaai17}
\usepackage{times}
\usepackage{helvet}
\usepackage{courier}
\usepackage{latexsym}
\usepackage{multirow}
\usepackage{pbox}
\usepackage[pdftex]{graphicx}
\usepackage{amsmath}
\usepackage{epstopdf}
\usepackage[export]{adjustbox}
\usepackage{subcaption}
\usepackage{flushend}
\usepackage{mathtools}
\renewcommand{\vec}[1]{\mathbf{#1}} 

\usepackage{tikz}
\usepackage{pgf}
\usepackage{tikz-qtree}
\usetikzlibrary{arrows,decorations.pathmorphing,backgrounds,positioning,fit,petri,shapes.misc, arrows.meta,shapes.geometric,decorations.markings}
\tikzset{->-/.style={decoration={
			markings,
			mark=at position 0.5 with {\arrow{>}}},postaction={decorate}}}
\tikzset{->--/.style={decoration={
			markings,
			mark=at position 0.4 with {\arrow{>}}},postaction={decorate}}}
    \newtheorem{theorem}{Theorem}

    \newenvironment{proof}[1][Proof]{\begin{trivlist}
    		\item[\hskip \labelsep {\bfseries #1}]}{\end{trivlist}}
    \newenvironment{definition}[1][Definition]{\begin{trivlist}
    		\item[\hskip \labelsep {\bfseries #1}]}{\end{trivlist}}

    \newcommand{\qed}{\nobreak \ifvmode \relax \else
    	\ifdim\lastskip<1.5em \hskip-\lastskip
    	\hskip1.5em plus0em minus0.5em \fi \nobreak
    	\vrule height0.65em width0.5em depth0.25em\fi}

\newcommand{\squishlist}{
 \begin{list}{$\bullet$}
  { \setlength{\itemsep}{0pt}
     \setlength{\parsep}{3pt}
     \setlength{\topsep}{3pt}
     \setlength{\partopsep}{0pt}
     \setlength{\leftmargin}{1.5em}
     \setlength{\labelwidth}{1em}
     \setlength{\labelsep}{0.5em} } }

\newcounter{Lcount}
\newcommand{\squishlisttwo}{
\begin{list}{\arabic{Lcount}. }
{ \usecounter{Lcount}
\setlength{\itemsep}{0pt}
\setlength{\parsep}{0pt}
\setlength{\topsep}{0pt}
\setlength{\partopsep}{0pt}
\setlength{\leftmargin}{2em}
\setlength{\labelwidth}{1.5em}
\setlength{\labelsep}{0.5em} } }

\newcommand{\squishend}{
\end{list} }

\begin{document}
%
\title{Efficient Dependency-Guided Named Entity Recognition}
\author{Zhanming Jie \and Aldrian Obaja Muis \and Wei Lu\\
 {Singapore University of Technology and Design}\\
 8 Somapah Road, Singapore, 487372\\
 \texttt{zhanming\_jie@mymail.sutd.edu.sg, \ \{aldrian\_muis,luwei\}@sutd.edu.sg}
}
\maketitle
\begin{abstract}
	

Named entity recognition (NER), which focuses on the extraction of semantically meaningful named entities and their semantic classes from  text,
serves as an indispensable component for several down-stream natural language processing (NLP) tasks such as relation extraction and event extraction.
Dependency trees, on the other hand, also convey crucial semantic-level information.
It has been shown previously that such information can be used to improve the performance of NER~\cite{sasano2008japanese,ling2012fine}.
In this work, we investigate on how to better utilize the structured information conveyed by dependency trees to improve the performance of NER.
Specifically, unlike existing approaches which only exploit dependency information for designing local features, we show that certain global structured information of the dependency trees can be exploited when building NER models where such information can provide guided learning and inference.
Through extensive experiments, we show that our proposed novel {\em dependency-guided} NER model performs competitively with  models based on conventional semi-Markov conditional random fields, while requiring significantly less running time.

%
\end{abstract}


\section{Introduction}
Named entity recognition (NER) is one of the most important tasks in the field of natural language processing (NLP). 
The task focuses on the extraction of named entities together with their semantic classes (such as {\em organization} or {\em person}) from text.
The extracted named entity information has been shown to be useful in various NLP tasks, including coreference resolution, question answering and relation extraction \cite{lao2010relational,lee2012joint,krishnamurthy2015learning}.

\begin{figure}[t!]
	\begin{subfigure}[b]{\linewidth}
		\begin{tikzpicture}[node distance=1.0mm and 1.0mm, >=Stealth, 	place/.style={draw=none, inner sep=0pt}]
		\node [](anode) [] {\footnotesize Foreign};
		\node [](bnode) [right=of anode, xshift=-2mm, yshift=0.28mm] {\footnotesize Minister};
		\node [](cnode) [right=of bnode, xshift=-2mm] {\footnotesize Shlomo};
		\node [](dnode) [right=of cnode, xshift=-2mm] {\footnotesize Ben};
		\node [](enode) [right=of dnode, xshift=3mm] {\footnotesize -};
		\node [](fnode) [right=of enode, xshift=3mm] {\footnotesize Ami};
		\node [](gnode) [right=of fnode, yshift=-0.57mm] {\footnotesize gave};
		\node [place](hnode) [right=of gnode, yshift=0.3mm] {\footnotesize a};
		\node [place](inode) [right=of hnode, yshift=0.3mm, xshift=2mm] {\footnotesize talk};
		
		\node [](at) [below=of anode,yshift=3.0mm]{\scriptsize NNP};
		\node [](bt) [below=of bnode,yshift=2.39mm] {\scriptsize NNP};
		\node [](ct) [below=of cnode,yshift=2.39mm] {\scriptsize NNP};
		\node [](dt) [below=of dnode,yshift=2.39mm]{\scriptsize NNP};
		\node [](et) [below=of enode,yshift=1.7mm]{\scriptsize HYPH};
		\node [](ft) [below=of fnode,yshift=2.39mm]{\scriptsize NNP};
		\node [](gt) [below=of gnode,yshift=2.95mm]{\scriptsize VBD};
		\node [place](ht) [below=of hnode]{\scriptsize DT};
		\node [place](it) [below=of inode,yshift=0.05mm]{\scriptsize NN};
		
		\node [](ae) [below=of at,yshift=2.45mm]{\small \textsc{o}};
		\node [](be) [below=of bt,yshift=2.5mm] {\small \textsc{o}};
		\node [](ce) [below=of ct,yshift=2.5mm] {\small \textsc{b-per}};
		\node [](de) [below=of dt,yshift=2.5mm]{\small \textsc{i-per}};
		\node [](ee) [below=of et,yshift=2.5mm]{\small \textsc{i-per}};
		\node [](fe) [below=of ft,yshift=2.5mm]{\small \textsc{i-per}};
		\node [](ge) [below=of gt,yshift=2.5mm]{\small \textsc{o}};
		\node [place](he) [below=of ht,yshift=0.1mm]{\small \textsc{o}};
		\node [place](ie) [below=of it,yshift=0.1mm]{\small \textsc{o}};
		
		\draw [line width=1pt, -{Stealth[length=3.5mm, open]},->] (bnode) to [out=110,in=40, looseness=1] node [above] {} (anode);
		\draw [line width=1pt, -{Stealth[length=3.5mm, open]},->] (fnode) to [out=120,in=60, looseness=0.8] node [above] {} (bnode);
		\draw [line width=1pt, -{Stealth[length=3.5mm, open]},->] (fnode) to [out=120,in=60, looseness=0.8] node [above] {} (cnode);
		\draw [line width=1pt, -{Stealth[length=3.5mm, open]},->] (fnode) to [out=120,in=45, looseness=1] node [above] {} (dnode);
		\draw [line width=1pt, -{Stealth[length=3.5mm, open]},->] (fnode) to [out=120,in=45, looseness=1] node [above] {} (enode);
		\draw [line width=1pt, -{Stealth[length=3.5mm, open]},->] (gnode) to [out=120,in=45, looseness=1.2] node [above] {} (fnode);
		\draw [line width=1pt, -{Stealth[length=3.5mm, open]},->] (gnode) to [out=70,in=100, looseness=1.2] node [above] {} (inode);
		\draw [line width=1pt, -{Stealth[length=3.5mm, open]},->] (inode) to [out=120,in=60, looseness=1.8] node [above] {} (hnode);
		\end{tikzpicture} 
	\end{subfigure}
	\begin{subfigure}[b]{\linewidth}
		\begin{tikzpicture}[node distance=1.0mm and 1.0mm, >=Stealth, 	place/.style={draw=none, inner sep=0pt}]
		\node [](anode) [] {\footnotesize The};
		\node [](bnode) [right=of anode] {\footnotesize House};
		\node [](cnode) [right=of bnode] {\footnotesize of};
		\node [](dnode) [right=of cnode, yshift=-0.3mm] {\footnotesize Representatives};
		\node [](enode) [right=of dnode, xshift=-2mm, yshift=0.2mm] {\footnotesize votes};
		\node [](fnode) [right=of enode, yshift=-0.2mm] {\footnotesize on};
		\node [place](gnode) [right=of fnode, xshift=0.5mm, yshift=0.35mm] {\footnotesize the};
		\node [place](hnode) [right=of gnode, xshift=1.5mm, yshift=-0.35mm] {\footnotesize measure};
		
		\node [](at) [below=of anode,yshift=2.3mm]{\scriptsize DT};
		\node [](bt) [below=of bnode,yshift=2.3mm] {\scriptsize NNP};
		\node [](ct) [below=of cnode,yshift=2.3mm] {\scriptsize IN};
		\node [](dt) [below=of dnode,yshift=2.9mm]{\scriptsize NNPS};
		\node [](et) [below=of enode,yshift=2.2mm]{\scriptsize VB};
		\node [](ft) [below=of fnode,yshift=2.2mm]{\scriptsize IN};
		\node [place](gt) [below=of gnode, yshift=-0.1mm]{\scriptsize DT};
		\node [place](ht) [below=of hnode, yshift=-0.1mm]{\scriptsize NN};
		
		\node [](ae) [below=of at,yshift=2.25mm]{\small \textsc{b-org}};
		\node [](be) [below=of bt,yshift=2.25mm] {\small \textsc{i-org}};
		\node [](ce) [,below=of ct,yshift=2.3mm] {\small \textsc{i-org}};
		\node [](de) [below=of dt,yshift=2.35mm]{\small \textsc{i-org}};
		\node [](ee) [below=of et,yshift=2.33mm]{\small \textsc{o}};
		\node [](fe) [below=of ft,yshift=2.33mm]{\small \textsc{o}};
		\node [place](ge) [below=of gt, yshift=-0.1mm]{\small \textsc{o}};
		\node [place](he) [below=of ht, yshift=-0.05mm]{\small \textsc{o}};
		
		\draw [line width=1pt, -{Stealth[length=3.5mm, open]},->] (bnode) to [out=120,in=40, looseness=1.5] node [above] {} (anode);
		\draw [line width=1pt, -{Stealth[length=3.5mm, open]},->] (enode) to [out=120,in=60, looseness=0.89] node [above] {} (bnode);
		\draw [line width=1pt, -{Stealth[length=3.5mm, open]},->] (bnode) to [out=50,in=130, looseness=1.4] node [above] {} (cnode);
		\draw [line width=1pt, -{Stealth[length=3.5mm, open]},->] (cnode) to [out=60,in=120, looseness=1] node [above] {} (dnode);
		\draw [line width=1pt, -{Stealth[length=3.5mm, open]},->] (enode) to [out=70,in=125, looseness=1.4] node [above] {} (fnode);
		\draw [line width=1pt, -{Stealth[length=3.5mm, open]},->] (fnode) to [out=70,in=100, looseness=1.2] node [above] {} (hnode);
		\draw [line width=1pt, -{Stealth[length=3.5mm, open]},->] (hnode) to [out=120,in=50, looseness=1.4] node [above] {} (gnode);
		\end{tikzpicture}
	\end{subfigure}
	\caption{Two sentences annotated with both dependency and named entity information. The edges on top of words represent the dependencies and the labels with IOB encoding are the entity types.}
	\label{fig:twoexample}
\end{figure}
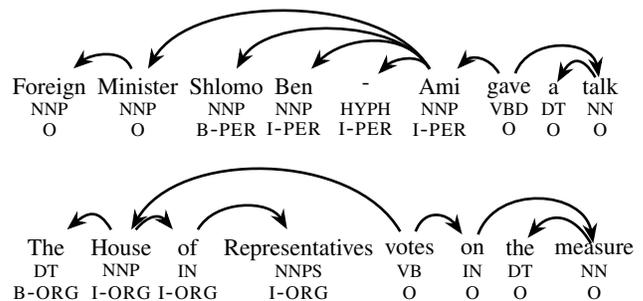

Dependency trees, on the other hand, were shown to be useful in several semantic processing tasks in NLP such as semantic parsing and question answering~\cite{poon2009unsupervised,liang2013learning}.
The dependency structures convey semantic-level information which was shown to be useful for the NER task.
Existing research efforts have exploited such dependency structured information by designing dependency-related local features  that can be used in the NER models~\cite{sasano2008japanese,ling2012fine,cucchiarelli2001unsupervised}. 
Figure \ref{fig:twoexample} shows two example phrases annotated with both dependency and named entity information. 
The local features are usually the head word and its part-of-speech tag at current position. 
For example, ``{\em Shlomo}'' with entity tag \textsc{b-per} in the first sentence has two local dependency features, head word ``{\em Ami}'' and  head tag ``{\em \textsc{nnp}}''. 
However, such a simple treatment of dependency structures largely ignores the global structured information conveyed by the dependency trees, which can be potentially useful in building NER models.

One key observation we can make in Figure 1 is that named entities are often covered by a single or multiple consecutive dependency arcs.
In the first example, the named entity ``{\em Shlomo Ben - Ami}'' of type \textsc{per} ({\em person}) is completely covered by the single dependency arc from ``{\em Ami}'' to ``{\em Shlomo}''. 
Similarly, the named entity ``{\em The House of Representatives}'' of type \textsc{org} ({\em organization}) in the second example is covered by multiple  arcs which are adjacent to each other. 
Such information can potentially be the global features we can obtain from the dependency trees. 
This leads to the following questions: 1) can such global structured information conveyed by dependency trees be exploited for improved NER, and 2) if so, how to build new NER models where such information can be explicitly incorporated?

With these two questions, in this paper we perform some investigations on how to better utilize the structured information conveyed by dependency trees for building novel models for improved named entity recognition.
The model assumes the availability of dependency trees before performing NER, which can be obtained from a dependency parser or given as part of the input.
Unlike existing approaches that only exploit dependency structures for encoding local features, the model is able to explicitly take into account the global  structured information conveyed by dependency trees when performing learning and inference.
We call our proposed NER model the {\em dependency-guided} model (\textsc{dgm}), and build it based on the conventional semi-Markov conditional random fields (semi-CRFs)~\cite{sarawagi2004semi}, a classic model used for information extraction.


Our main contributions can be summarized as follows:
\begin{itemize}
	\item We present a novel model that is able to explicitly exploit the global structured information conveyed by dependency trees, showing that such information can be effectively integrated into the process of performing NER. To the best of our knowledge, this is the first  work that exploits such information for NER.
	\item Theoretically, we show through average-case time complexity analysis that our model has the same time complexity as that of the linear-chain CRFs, and is better than that of the semi-Markov CRFs.
	\item Empirically, we demonstrate the benefits of our approach through extensive experiments on benchmark datasets.
	We show that the resulting  model leads to NER results that are competitive with the baseline approach based on semi-Markov CRFs, while requiring significantly less running time.
\end{itemize}

Furthermore, although in this paper we  focus on the task of using the structured information conveyed by dependency trees for improved NER, the underlying idea is general, which we believe can be applied to other tasks that involve building pipeline or joint models for structured prediction. 
Potentially if we can find the relationship between different NLP tasks, we can build an efficient model for a specific task while using information from other tasks without making the model more complex. 


\section{Related Work}
Named entity recognition has a long history in the field of natural language processing. 
One standard approach to NER is to regard the problem as a sequence labeling problem,
where each word is assigned a tag, indicating whether the word belongs to part of any named entity or appears outside of all entities.
Previous approaches used sequence labeling models such as hidden Markov models (HMMs)~\cite{zhou2002named}, maximum entropy Markov models (MEMMs)~\cite{mccallum2000maximum}, as well as linear-chain~\cite{finkel2005incorporating} and semi-Markov conditional random fields (CRFs/semi-CRFs)~\cite{sarawagi2004semi}. 
\citeauthor{muis2016weak} \shortcite{muis2016weak} proposed a weak semi-CRFs model which has a lower complexity than the conventional semi-CRFs model while still having a higher complexity than the linear-chain CRFs model. 
Our model is proved to have the same time complexity as linear-chain CRFs model in the average case. 
The quality of the CRFs model typically depends on the features that are used.
While most research efforts exploited standard word-level features~\cite{ratinov2009design}, more sophisticated features can also be used. 
\citeauthor{ling2012fine} \shortcite{ling2012fine} showed that using syntactic-level features from dependency structures in a CRFs-based model can lead to improved NER performance.
Such dependency structures were also used in the work by \citeauthor{liu2010recognizing} \shortcite{liu2010recognizing},
where the authors utilized such structures for building a skip-chain variant of the original CRFs model.
This shows that some simple structured information conveyed by dependency trees can be exploited for improved NER. 
In their skip-chain CRFs model, they simply added certain dependency arcs as additional dependencies in the graphical model, resulting in loopy structures.
However, such a model did not explicitly explore the relation between entities and global structured information of the dependency trees.
The authors also showed that such a model does not outperform a simpler approach that adds additional dependencies between similar words only on top of the original CRFs model.
%
In this work, we also focus on utilizing dependency structures for improving NER.
Unlike previous approaches, we focus on exploiting the global structured information conveyed by dependency trees to improve the NER process. Comparing with the semi-CRFs model, our  model is not only able to perform competitively in terms of performance, but also more {\em efficient} in terms of running time.

There are also some existing works that focus on improving the efficiency of NER and other information extraction models.
For example, \citeauthor{okanohara2006improving} \shortcite{okanohara2006improving} used a separate naive Bayes classifier to filter some entities during training and inference in their semi-CRFs based model.
While the filtering process was used to reduce the computational cost of the semi-CRFs model, the model still needs to enumerate all the possible chunks. 
\citeauthor{yang2012extracting} \shortcite{yang2012extracting} extended the original semi-CRFs for extracting opinion expressions and used the constituency parse tree information to avoid constructing implausible segments. 
\citeauthor{lu2015joint} \shortcite{lu2015joint} proposed an efficient and scalable model using hypergraph which can handle overlapping entities. 
\citeauthor{muis2016learning} \shortcite{muis2016learning} extended the hypergraph representation to recognize both contiguous and discontiguous entities. 


\section{Background}
Before we introduce our  models, we would like to have a brief discussion on the relevant background. 
Specifically, in this section, we  review the two classic models that are commonly used for named entity recognition, namely the linear-chain conditional random fields and the semi-Markov conditional random fields models.

\subsection{Linear-chain CRFs}
Conditional random fields, or CRFs~\cite{lafferty2001conditional} is a popular model for structured prediction, which has been widely used in various natural language processing problems, including named entity recognition~\cite{mccallum2003early} and semantic role labeling~\cite{cohn2005semantic}. 

We focus our discussions on the linear-chain CRFs in this section.
The probability of predicting a possible output sequence $\vec{y}$ ({\em e.g.}, a named entity label sequence in our case) given an input $\vec{x}$ ({\em e.g.}, a sentence) is defined as:
\begin{equation}
p(\vec{y}|\vec{x}) = \frac{\exp(\vec{\vec{w}^{T}\vec{f}(\vec{x},\vec{y}) })}{Z(\vec{x})}
\end{equation}
where $\vec{f}(\vec{x},\vec{y})$ is the feature vector defined over $(\vec{x},\vec{y})$ tuple, $\vec{w}$ is the weight vector consisting of parameters used for the model, and $Z(\vec{x})$ is the partition function used for normalization, which is defined as follows:
\begin{equation}
Z(\vec{x}) = \sum_{\vec{y}}
\exp
(\vec{w}^{T}\vec{f}(\vec{x}, \vec{y}))
\end{equation}

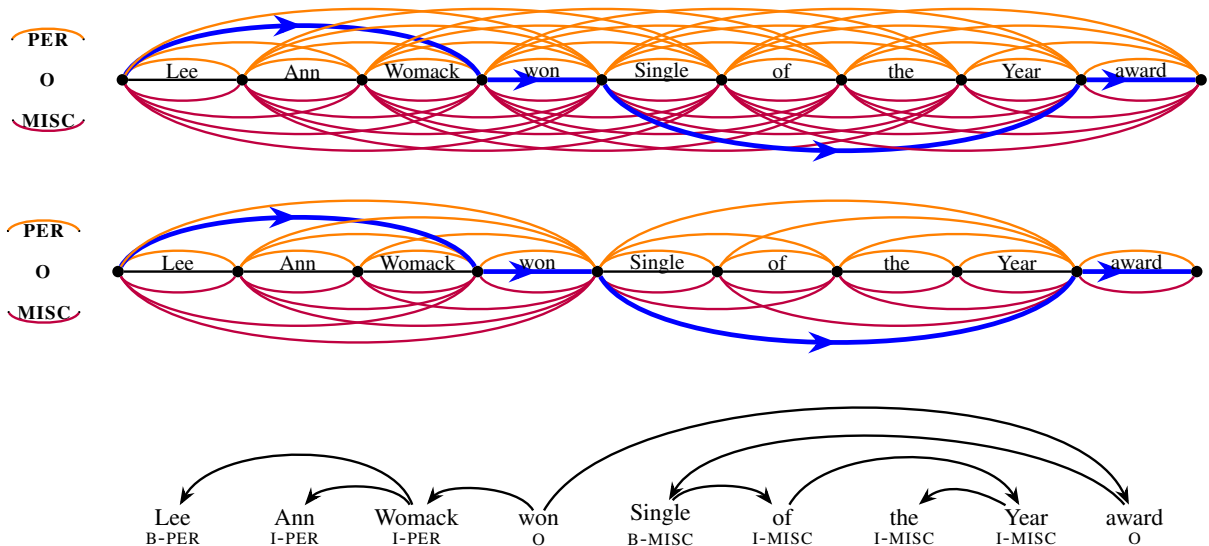
\begin{figure*}[ht!]
	\centering
	\begin{adjustbox}{minipage=\linewidth,scale=0.9}
		\begin{subfigure}[b]{\columnwidth}
			\centering
			\begin{tikzpicture}[node distance=5mm and 16mm, 
			>=Stealth, inner sep=0mm,
			place/.style={circle, draw, thick, fill=black, inner sep=0.5mm}, smalldot/.style={circle,draw, fill=black,scale=0.5}]			
			\node at (0, 0) [place] (p0){};
			\node [left = of p0, xshift = 9mm, yshift = 6mm](per1) {\textbf{\textsc{per}}};
			\node [left = of per1, xshift = 14mm, smalldot](pr1left) {};
			\node [right = of per1, xshift = -14mm, smalldot](pr1right) {};
			\draw [line width=1pt, -{Stealth[length=3.5mm, open]},-,orange] (pr1left) to [out=60,in=120, looseness=0.55] node [above] {} (pr1right);
			\node [left = of p0, xshift = 7mm, yshift = 0mm](o1) {\textbf{\textsc{o}}};
			\node [left = of p0, xshift = 10mm, yshift = -6mm](misc1) {\textbf{\textsc{misc}}};
			\node [left = of misc1, xshift = 15mm, smalldot](misc1left) {};
			\node [right = of misc1, xshift = -15mm, smalldot](misc1right) {};
			\draw [line width=1pt, -{Stealth[length=3.5mm, open]},-,purple] (misc1left) to [out=-60,in=-120, looseness=0.55] node [above] {} (misc1right);
			\node at (0, 0) [place, right = of p0] (p1){};
			\node at (0, 0) [place, right = of p1] (p2){};
			\node at (0, 0) [place, right = of p2] (p3){};
			\node at (0, 0) [place, right = of p3] (p4){};
			\node at (0, 0) [place, right = of p4] (p5){};
			\node at (0, 0) [place, right = of p5] (p6){};
			\node at (0, 0) [place, right = of p6] (p7){};
			\node at (0, 0) [place, right = of p7] (p8){};
			\node at (0, 0) [place, right = of p8] (p9){};
			
			\draw [line width=1pt, -{Stealth[length=3.5mm, open]},-,orange] (p0) to [out=60,in=120, looseness=0.55] node [above] {} (p1);
			\draw [line width=1pt, -{Stealth[length=3.5mm, open]},-,orange] (p1) to [out=60,in=120, looseness=0.55] node [above] {} (p2);
			\draw [line width=1pt, -{Stealth[length=3.5mm, open]},-,orange] (p2) to [out=60,in=120, looseness=0.55] node [above] {} (p3);
			\draw [line width=1pt, -{Stealth[length=3.5mm, open]},-,orange] (p3) to [out=60,in=120, looseness=0.55] node [above] {} (p4);
			\draw [line width=1pt, -{Stealth[length=3.5mm, open]},-,orange] (p4) to [out=60,in=120, looseness=0.55] node [above] {} (p5);
			\draw [line width=1pt, -{Stealth[length=3.5mm, open]},-,orange] (p5) to [out=60,in=120, looseness=0.55] node [above] {} (p6);
			\draw [line width=1pt, -{Stealth[length=3.5mm, open]},-,orange] (p6) to [out=60,in=120, looseness=0.55] node [above] {} (p7);
			\draw [line width=1pt, -{Stealth[length=3.5mm, open]},-,orange] (p7) to [out=60,in=120, looseness=0.55] node [above] {} (p8);
			\draw [line width=1pt, -{Stealth[length=3.5mm, open]},-,orange] (p8) to [out=60,in=120, looseness=0.55] node [above] {} (p9);
			
			\draw [line width=1pt, -{Stealth[length=3.5mm, open]},-,purple] (p0) to [out=-60,in=-120, looseness=0.55] node [above] {} (p1);
			\draw [line width=1pt, -{Stealth[length=3.5mm, open]},-,purple] (p1) to [out=-60,in=-120, looseness=0.55] node [above] {} (p2);
			\draw [line width=1pt, -{Stealth[length=3.5mm, open]},-,purple] (p2) to [out=-60,in=-120, looseness=0.55] node [above] {} (p3);
			\draw [line width=1pt, -{Stealth[length=3.5mm, open]},-,purple] (p3) to [out=-60,in=-120, looseness=0.55] node [above] {} (p4);
			\draw [line width=1pt, -{Stealth[length=3.5mm, open]},-,purple] (p4) to [out=-60,in=-120, looseness=0.55] node [above] {} (p5);
			\draw [line width=1pt, -{Stealth[length=3.5mm, open]},-,purple] (p5) to [out=-60,in=-120, looseness=0.55] node [above] {} (p6);
			\draw [line width=1pt, -{Stealth[length=3.5mm, open]},-,purple] (p6) to [out=-60,in=-120, looseness=0.55] node [above] {} (p7);
			\draw [line width=1pt, -{Stealth[length=3.5mm, open]},-,purple] (p7) to [out=-60,in=-120, looseness=0.55] node [above] {} (p8);
			\draw [line width=1pt, -{Stealth[length=3.5mm, open]},-,purple] (p8) to [out=-60,in=-120, looseness=0.55] node [above] {} (p9);
			
			\draw [line width=1pt,-] (p0) to [] node [above] {\small Lee} (p1);
			\draw [line width=1pt,-] (p1) to [] node [above] {\small Ann} (p2);
			\draw [line width=1pt,-] (p2) to [] node [above] {\small Womack} (p3);
			\draw [line width=2pt,->-, blue] (p3) to [] node [above,black] {\small won} (p4);
			\draw [line width=1pt,-] (p4) to [] node [above,yshift=-0.5mm] {\small Single} (p5);
			\draw [line width=1pt,-] (p5) to [] node [above] {\small of} (p6);
			\draw [line width=1pt,-] (p6) to [] node [above] {\small the} (p7);
			\draw [line width=1pt,-] (p7) to [] node [above] {\small Year} (p8);
			\draw [line width=2pt,->--, blue] (p8) to [] node [above,black] {\small award} (p9);
			
			\draw [line width=1pt, -{Stealth[length=3.5mm, open]},-,orange] (p0) to [out=60,in=120, looseness=0.55] node [above] {} (p2);
			\draw [line width=1pt, -{Stealth[length=3.5mm, open]},-,orange] (p1) to [out=60,in=120, looseness=0.55] node [above] {} (p3);
			\draw [line width=1pt, -{Stealth[length=3.5mm, open]},-,orange] (p2) to [out=60,in=120, looseness=0.55] node [above] {} (p4);
			\draw [line width=1pt, -{Stealth[length=3.5mm, open]},-,orange] (p3) to [out=60,in=120, looseness=0.55] node [above] {} (p5);
			\draw [line width=1pt, -{Stealth[length=3.5mm, open]},-,orange] (p4) to [out=60,in=120, looseness=0.55] node [above] {} (p6);
			\draw [line width=1pt, -{Stealth[length=3.5mm, open]},-,orange] (p5) to [out=60,in=120, looseness=0.55] node [above] {} (p7);
			\draw [line width=1pt, -{Stealth[length=3.5mm, open]},-,orange] (p6) to [out=60,in=120, looseness=0.55] node [above] {} (p8);
			\draw [line width=1pt, -{Stealth[length=3.5mm, open]},-,orange] (p7) to [out=60,in=120, looseness=0.55] node [above] {} (p9);
			
			\draw [line width=1pt, -{Stealth[length=3.5mm, open]},-,purple] (p0) to [out=-60,in=-120, looseness=0.55] node [above] {} (p2);
			\draw [line width=1pt, -{Stealth[length=3.5mm, open]},-,purple] (p1) to [out=-60,in=-120, looseness=0.55] node [above] {} (p3);
			\draw [line width=1pt, -{Stealth[length=3.5mm, open]},-,purple] (p2) to [out=-60,in=-120, looseness=0.55] node [above] {} (p4);
			\draw [line width=1pt, -{Stealth[length=3.5mm, open]},-,purple] (p3) to [out=-60,in=-120, looseness=0.55] node [above] {} (p5);
			\draw [line width=1pt, -{Stealth[length=3.5mm, open]},-,purple] (p4) to [out=-60,in=-120, looseness=0.55] node [above] {} (p6);
			\draw [line width=1pt, -{Stealth[length=3.5mm, open]},-,purple] (p5) to [out=-60,in=-120, looseness=0.55] node [above] {} (p7);
			\draw [line width=1pt, -{Stealth[length=3.5mm, open]},-,purple] (p6) to [out=-60,in=-120, looseness=0.55] node [above] {} (p8);
			\draw [line width=1pt, -{Stealth[length=3.5mm, open]},-,purple] (p7) to [out=-60,in=-120, looseness=0.55] node [above] {} (p9);
			
			\draw [line width=2pt, ->-, blue] (p0) to [out=60,in=120, looseness=0.55] node [above] {} (p3);
			\draw [line width=1pt, -{Stealth[length=3.5mm, open]},-,orange] (p1) to [out=60,in=120, looseness=0.55] node [above] {} (p4);
			\draw [line width=1pt, -{Stealth[length=3.5mm, open]},-,orange] (p2) to [out=60,in=120, looseness=0.55] node [above] {} (p5);
			\draw [line width=1pt, -{Stealth[length=3.5mm, open]},-,orange] (p3) to [out=60,in=120, looseness=0.55] node [above] {} (p6);
			\draw [line width=1pt, -{Stealth[length=3.5mm, open]},-,orange] (p4) to [out=60,in=120, looseness=0.55] node [above] {} (p7);
			\draw [line width=1pt, -{Stealth[length=3.5mm, open]},-,orange] (p5) to [out=60,in=120, looseness=0.55] node [above] {} (p8);
			\draw [line width=1pt, -{Stealth[length=3.5mm, open]},-,orange] (p6) to [out=60,in=120, looseness=0.55] node [above] {} (p9);
			
			\draw [line width=1pt, -{Stealth[length=3.5mm, open]},-,purple] (p0) to [out=-60,in=-120, looseness=0.55] node [above] {} (p3);
			\draw [line width=1pt, -{Stealth[length=3.5mm, open]},-,purple] (p1) to [out=-60,in=-120, looseness=0.55] node [above] {} (p4);
			\draw [line width=1pt, -{Stealth[length=3.5mm, open]},-,purple] (p2) to [out=-60,in=-120, looseness=0.55] node [above] {} (p5);
			\draw [line width=1pt, -{Stealth[length=3.5mm, open]},-,purple] (p3) to [out=-60,in=-120, looseness=0.55] node [above] {} (p6);
			\draw [line width=1pt, -{Stealth[length=3.5mm, open]},-,purple] (p4) to [out=-60,in=-120, looseness=0.55] node [above] {} (p7);
			\draw [line width=1pt, -{Stealth[length=3.5mm, open]},-,purple] (p5) to [out=-60,in=-120, looseness=0.55] node [above] {} (p8);
			\draw [line width=1pt, -{Stealth[length=3.5mm, open]},-,purple] (p6) to [out=-60,in=-120, looseness=0.55] node [above] {} (p9);
			
			\draw [line width=1pt, -{Stealth[length=3.5mm, open]},-,orange] (p0) to [out=60,in=120, looseness=0.55] node [above] {} (p4);
			\draw [line width=1pt, -{Stealth[length=3.5mm, open]},-,orange] (p1) to [out=60,in=120, looseness=0.55] node [above] {} (p5);
			\draw [line width=1pt, -{Stealth[length=3.5mm, open]},-,orange] (p2) to [out=60,in=120, looseness=0.55] node [above] {} (p6);
			\draw [line width=1pt, -{Stealth[length=3.5mm, open]},-,orange] (p3) to [out=60,in=120, looseness=0.55] node [above] {} (p7);
			\draw [line width=1pt, -{Stealth[length=3.5mm, open]},-,orange] (p4) to [out=60,in=120, looseness=0.55] node [above] {} (p8);
			\draw [line width=1pt, -{Stealth[length=3.5mm, open]},-,orange] (p5) to [out=60,in=120, looseness=0.55] node [above] {} (p9);
			
			\draw [line width=1pt, -{Stealth[length=3.5mm, open]},-,purple] (p0) to [out=-60,in=-120, looseness=0.55] node [above] {} (p4);
			\draw [line width=1pt, -{Stealth[length=3.5mm, open]},-,purple] (p1) to [out=-60,in=-120, looseness=0.55] node [above] {} (p5);
			\draw [line width=1pt, -{Stealth[length=3.5mm, open]},-,purple] (p2) to [out=-60,in=-120, looseness=0.55] node [above] {} (p6);
			\draw [line width=1pt, -{Stealth[length=3.5mm, open]},-,purple] (p3) to [out=-60,in=-120, looseness=0.55] node [above] {} (p7);
			\draw [line width=2pt, ->-, blue] (p4) to [out=-60,in=-120, looseness=0.55] node [above] {} (p8);
			\draw [line width=1pt, -{Stealth[length=3.5mm, open]},-,purple] (p5) to [out=-60,in=-120, looseness=0.55] node [above] {} (p9);
			\end{tikzpicture}
		\end{subfigure} 
		\begin{subfigure}[b]{\columnwidth}
			\begin{tikzpicture}[node distance=5mm and 16mm, >=Stealth, inner sep=0mm,
			place/.style={circle, draw, thick, fill=black, inner sep=0.5mm}, smalldot/.style={circle,draw, fill=black,scale=0.5}]
			  
			\node at (0, 0) [place] (p0){};
			\node [left = of p0, xshift = 9mm, yshift = 6mm](per1) {\textbf{\textsc{per}}};
			\node [left = of per1, xshift = 14mm, smalldot](pr1left) {};
			\node [right = of per1, xshift = -14mm, smalldot](pr1right) {};
			\draw [line width=1pt, -{Stealth[length=3.5mm, open]},-,orange] (pr1left) to [out=60,in=120, looseness=0.55] node [above] {} (pr1right);
			\node [left = of p0, xshift = 7mm, yshift = 0mm](o1) {\textbf{\textsc{o}}};
			\node [left = of p0, xshift = 10mm, yshift = -6mm](misc1) {\textbf{\textsc{misc}}};
			\node [left = of misc1, xshift = 15mm, smalldot](misc1left) {};
			\node [right = of misc1, xshift = -15mm, smalldot](misc1right) {};
			\draw [line width=1pt, -{Stealth[length=3.5mm, open]},-,purple] (misc1left) to [out=-60,in=-120, looseness=0.55] node [above] {} (misc1right);
			\node at (0, 0) [place, right = of p0] (p1){};
			\node at (0, 0) [place, right = of p1] (p2){};
			\node at (0, 0) [place, right = of p2] (p3){};
			\node at (0, 0) [place, right = of p3] (p4){};
			\node at (0, 0) [place, right = of p4] (p5){};
			\node at (0, 0) [place, right = of p5] (p6){};
			\node at (0, 0) [place, right = of p6] (p7){};
			\node at (0, 0) [place, right = of p7] (p8){};
			\node at (0, 0) [place, right = of p8] (p9){};
			
			\draw [line width=1pt, -{Stealth[length=3.5mm, open]},-,orange] (p0) to [out=60,in=120, looseness=0.55] node [above] {} (p1);
			\draw [line width=1pt, -{Stealth[length=3.5mm, open]},-,orange] (p1) to [out=60,in=120, looseness=0.55] node [above] {} (p2);
			\draw [line width=1pt, -{Stealth[length=3.5mm, open]},-,orange] (p2) to [out=60,in=120, looseness=0.55] node [above] {} (p3);
			\draw [line width=1pt, -{Stealth[length=3.5mm, open]},-,orange] (p3) to [out=60,in=120, looseness=0.55] node [above] {} (p4);
			\draw [line width=1pt, -{Stealth[length=3.5mm, open]},-,orange] (p4) to [out=60,in=120, looseness=0.55] node [above] {} (p5);
			\draw [line width=1pt, -{Stealth[length=3.5mm, open]},-,orange] (p5) to [out=60,in=120, looseness=0.55] node [above] {} (p6);
			\draw [line width=1pt, -{Stealth[length=3.5mm, open]},-,orange] (p6) to [out=60,in=120, looseness=0.55] node [above] {} (p7);
			\draw [line width=1pt, -{Stealth[length=3.5mm, open]},-,orange] (p7) to [out=60,in=120, looseness=0.55] node [above] {} (p8);
			\draw [line width=1pt, -{Stealth[length=3.5mm, open]},-,orange] (p8) to [out=60,in=120, looseness=0.55] node [above] {} (p9);
			
			\draw [line width=1pt, -{Stealth[length=3.5mm, open]},-,purple] (p0) to [out=-60,in=-120, looseness=0.55] node [above] {} (p1);
			\draw [line width=1pt, -{Stealth[length=3.5mm, open]},-,purple] (p1) to [out=-60,in=-120, looseness=0.55] node [above] {} (p2);
			\draw [line width=1pt, -{Stealth[length=3.5mm, open]},-,purple] (p2) to [out=-60,in=-120, looseness=0.55] node [above] {} (p3);
			\draw [line width=1pt, -{Stealth[length=3.5mm, open]},-,purple] (p3) to [out=-60,in=-120, looseness=0.55] node [above] {} (p4);
			\draw [line width=1pt, -{Stealth[length=3.5mm, open]},-,purple] (p4) to [out=-60,in=-120, looseness=0.55] node [above] {} (p5);
			\draw [line width=1pt, -{Stealth[length=3.5mm, open]},-,purple] (p5) to [out=-60,in=-120, looseness=0.55] node [above] {} (p6);
			\draw [line width=1pt, -{Stealth[length=3.5mm, open]},-,purple] (p6) to [out=-60,in=-120, looseness=0.55] node [above] {} (p7);
			\draw [line width=1pt, -{Stealth[length=3.5mm, open]},-,purple] (p7) to [out=-60,in=-120, looseness=0.55] node [above] {} (p8);
			\draw [line width=1pt, -{Stealth[length=3.5mm, open]},-,purple] (p8) to [out=-60,in=-120, looseness=0.55] node [above] {} (p9);
			
			\draw [line width=1pt,-] (p0) to [] node [above] {\small Lee} (p1);
			\draw [line width=1pt,-] (p1) to [] node [above] {\small Ann} (p2);
			\draw [line width=1pt,-] (p2) to [] node [above] {\small Womack} (p3);
			\draw [line width=2pt,->-,blue] (p3) to [] node [above,black] {\small won} (p4);
			\draw [line width=1pt,-] (p4) to [] node [above, yshift=-0.5mm] {\small Single} (p5);
			\draw [line width=1pt,-] (p5) to [] node [above] {\small of} (p6);
			\draw [line width=1pt,-] (p6) to [] node [above] {\small the} (p7);
			\draw [line width=1pt,-] (p7) to [] node [above] {\small Year} (p8);
			\draw [line width=2pt,->--, blue] (p8) to [] node [above,black] {\small award} (p9);
			
			\draw [line width=1pt, -{Stealth[length=3.5mm, open]},-,orange] (p1) to [out=60,in=120, looseness=0.55] node [above] {} (p3);
			\draw [line width=1pt, -{Stealth[length=3.5mm, open]},-,orange] (p2) to [out=60,in=120, looseness=0.55] node [above] {} (p4);
			\draw [line width=1pt, -{Stealth[length=3.5mm, open]},-,orange] (p4) to [out=60,in=120, looseness=0.55] node [above] {} (p6);
			\draw [line width=1pt, -{Stealth[length=3.5mm, open]},-,orange] (p6) to [out=60,in=120, looseness=0.55] node [above] {} (p8);
			
			\draw [line width=1pt, -{Stealth[length=3.5mm, open]},-,purple] (p1) to [out=-60,in=-120, looseness=0.55] node [above] {} (p3);
			\draw [line width=1pt, -{Stealth[length=3.5mm, open]},-,purple] (p2) to [out=-60,in=-120, looseness=0.55] node [above] {} (p4);
			\draw [line width=1pt, -{Stealth[length=3.5mm, open]},-,purple] (p4) to [out=-60,in=-120, looseness=0.55] node [above] {} (p6);
			\draw [line width=1pt, -{Stealth[length=3.5mm, open]},-,purple] (p6) to [out=-60,in=-120, looseness=0.55] node [above] {} (p8);

			\draw [line width=2pt, ->-,blue] (p0) to [out=60,in=120, looseness=0.55] node [above] {} (p3);
			\draw [line width=1pt, -{Stealth[length=3.5mm, open]},-,orange] (p1) to [out=60,in=120, looseness=0.55] node [above] {} (p4);
			\draw [line width=1pt, -{Stealth[length=3.5mm, open]},-,orange] (p5) to [out=60,in=120, looseness=0.55] node [above] {} (p8);

			\draw [line width=1pt, -{Stealth[length=3.5mm, open]},-,purple] (p0) to [out=-60,in=-120, looseness=0.55] node [above] {} (p3);
			\draw [line width=1pt, -{Stealth[length=3.5mm, open]},-,purple] (p1) to [out=-60,in=-120, looseness=0.55] node [above] {} (p4);
			\draw [line width=1pt, -{Stealth[length=3.5mm, open]},-,purple] (p5) to [out=-60,in=-120, looseness=0.55] node [above] {} (p8);

			\draw [line width=1pt, -{Stealth[length=3.5mm, open]},-,orange] (p0) to [out=60,in=120, looseness=0.55] node [above] {} (p4);
			\draw [line width=1pt, -{Stealth[length=3.5mm, open]},-,orange] (p4) to [out=60,in=120, looseness=0.55] node [above] {} (p8);

			\draw [line width=1pt, -{Stealth[length=3.5mm, open]},-,purple] (p0) to [out=-60,in=-120, looseness=0.55] node [above] {} (p4);
			\draw [line width=2pt,->-, blue] (p4) to [out=-60,in=-120, looseness=0.55] node [above] {} (p8);

			\end{tikzpicture}
		\end{subfigure}
		\begin{subfigure}[b]{\columnwidth}
			\begin{tikzpicture}[node distance=3mm and 16mm, 
			>=Stealth, inner sep=0.5mm,
			place/.style={circle, draw, thick, fill=black}]
			\node at (0mm, -25mm) [xshift = 0mm, yshift = 8mm](per1) {};
			\node at (24mm, -25mm) [label=below:\small \textsc{b-per},  yshift = -0.05mm] (p0w) {Lee};
			\node at (42mm, -25mm) [label=below:\small \textsc{i-per},  yshift = -0.05mm] (p1w) {Ann};
			\node at (60mm, -25mm) [label=below:\small \textsc{i-per}] (p2w) {Womack};
			\node at (78mm, -25.6mm) [label=below:\small \textsc{o},  yshift = 0.15mm] (p3w) {won};
			\node at (96mm, -24.9mm) [label=below:\small \textsc{b-misc},  yshift = 0.2mm] (p4w) {Single};
			\node at (114mm, -25mm) [label=below:\small \textsc{i-misc}] (p5w) {of};
			\node at (132mm, -25mm) [label=below:\small \textsc{i-misc}] (p6w) {the};
			\node at (150mm, -25mm) [label=below:\small \textsc{i-misc},  yshift = -0.05mm] (p7w) {Year};
			\node at (166mm, -25mm) [label=below:\small \textsc{o}] (p8w) {award};
			\draw [line width=1pt, -{Stealth[length=3.5mm, open]},->] (p2w) to [out=110,in=60, looseness=0.8] node [above] {} (p0w);
			\draw [line width=1pt, -{Stealth[length=3.5mm, open]},->] (p2w) to [out=120,in=50, looseness=0.7] node [above] {} (p1w);
			\draw [line width=1pt, -{Stealth[length=3.5mm, open]},->] (p3w) to [out=120,in=50, looseness=0.7] node [above] {} (p2w);
			\draw [line width=1pt, -{Stealth[length=3.5mm, open]},->] (p3w) to [out=60,in=115, looseness=0.65] node [above] {} (p8w);
			\draw [line width=1pt, -{Stealth[length=3.5mm, open]},->] (p4w) to [out=50,in=130, looseness=0.7] node [above] {} (p5w);
			\draw [line width=1pt, -{Stealth[length=3.5mm, open]},->] (p5w) to [out=60,in=120, looseness=0.8] node [above] {} (p7w);
			\draw [line width=1pt, -{Stealth[length=3.5mm, open]},->] (p7w) to [out=150,in=40, looseness=1] node [above] {} (p6w);
			\draw [line width=1pt, -{Stealth[length=3.5mm, open]},->] (p8w) to [out=130,in=60, looseness=0.6] node [above] {} (p4w);
			\end{tikzpicture}
			
		\end{subfigure}

	\end{adjustbox}
	\caption{Illustrations of possible combinations of entities for the conventional semi-CRFs model (top) and our \textsc{dgm} model (middle), as well as the example sentence with its dependency structure (bottom).}
	\label{fig:graphexample}
\end{figure*}

In a (first-order) linear-chain CRF, the partition function for a input of length $n$ can also be written as follows:
\begin{equation}
Z(\mathbf{x}) = \sum_{\mathbf{y}}\exp\smashoperator{\sum_{(y',y,i)\in\mathcal{E}(\mathbf{x}, \mathbf{y})}}\ \mathbf{w}^T\mathbf{f}(\vec{x}, y',y,i)
\end{equation}
where $\mathcal{E}(\vec{x}, \vec{y})$ is the set of edges which defines the input $\vec{x}$ labeled with the label sequence $\vec{y}$ and $\vec{f}(\vec{x},y', y,i)$ is a local feature vector defined at the $i$-th position of the input sequence.
$T$ is the set of all output labels.
The above function can be computed efficiently using dynamic programming.
It can be seen that the time complexity of a linear-chain CRFs model is $\mathcal{O}(n|T|^{2})$.

We aim to minimize the negative joint log-likelihood with $\textit{L}_{2}$ regularization for our dataset:
\begin{equation}
\begin{split}
\mathcal{L}(\vec{w}) = & \sum_{i}\log\sum_{\substack{\vec{y}{'} } }\exp(\vec{w}^{T}\vec{f}(\vec{x}_{i},\vec{y}{'})) \\
& - \sum_{i}\vec{w}^{T}\vec{f}(\vec{x}_{i},\vec{y}_{i}) + \lambda \vec{w}^{T}\vec{w}
\end{split}
\end{equation} 
where $(\vec{x}_{i},\vec{y}_{i})$ is the $i$-th training instance and $\lambda$ is the $L_{2}$ regularization coefficient.

The objective function is convex and we can make use of the L-BFGS \cite{byrd1995limited} algorithm to optimize it.
The partial derivative of $\mathcal{L}$ with respect to each parameter $w_{k}$ is:
\begin{equation}
\frac{\partial\mathcal{L}}{\partial w_{k}}  = \sum_i\Big(\mathbf{E}_{p(\vec{y}{'}|\vec{x}_{i})} \left [f_{k}(\vec{x}_{i}, \vec{y}{'})    \right ] -  f_{k}(\vec{x}_{i}, \vec{y}_{i})\Big) + 2\lambda w_{k}\nonumber
\end{equation}

\subsection{Semi-Markov CRFs}
The semi-Markov conditional random fields, or semi-CRFs~\cite{sarawagi2004semi} is an extension of the standard linear-chain CRFs.
Different from linear-chain CRFs, which makes use of simple first-order Markovian assumptions, the semi-CRFs
assumes that the transitions between words inside a span ({\em e.g.}, an entity consisting of multiple words) can be non-Markovian.
Such an assumption allows more flexible non-Markovian features to be defined.
The resulting model was shown to be more effective than its linear-chain counterpart in information extraction tasks.


The partition function in this setting becomes:
\begin{equation}
Z(\vec{x}) = 
\sum_\vec{y}
\exp \!\!\!\!\! \!\!\!\!\!
\sum_{(y^\prime, y, i-l,i) \in \mathcal{E}(\vec{x}, \vec{y})}
\!\!\!\! \!\!\!\!\!
\vec{w}^{T}\vec{f}(\vec{x}, y^\prime, y, i-l,i)
\end{equation}
where $\vec{f}(\vec{x},y', y, i-l, i)$ represents the local feature vector at position $i$ with a span of size $l$. In this case, the time complexity becomes $\mathcal{O} (nL|T|^{2})$.
The value $L$ is the maximal length of the spans the model can handle.

Consider the sentence ``{\em Lee Ann Womack won Single of the Year award}''.
The upper portion of Figure \ref{fig:graphexample} shows all the possible structures that are considered by the partition function with $L=4$.
These structures essentially form a compact lattice representation showing all the possible combinations of entities (with length restriction)
that can ever appear in the given sentence.
Each orange and red edge is used to represent an entity of type \textsc{per} and \textsc{misc} respectively.
The black lines are used to indicate words that are not part of any entity.
The directed path highlighted in bold blue  corresponds to the correct  entity information associated with the phrase, where ``{\em Lee Ann Womack}'' is an entity of type \textsc{per}, and ``{\em Single of the Year}'' is another entity of  type \textsc{misc} ({\em miscellaneous}).

\section{Our Model}

The primary assumption we would like to make  is that the dependency trees of sentences are available when performing the  named entity recognition task.


\subsection{NER with Dependency Features}

One  approach to exploit information from dependency structures is to design local features based on such dependency structures and make use of conventional models such as semi-CRFs for performing NER.
Previous research efforts have shown that such features extracted from dependency structures can be helpful when performing the entity recognition task~\cite{sasano2008japanese,ling2012fine,cucchiarelli2001unsupervised}.

In Figure~\ref{fig:graphexample}, we have already used a graphical illustration to show the possible combinations of entities that can ever appear in the given sentence.
Each edge in the figure corresponds to one entity (or a single word that is outside of any entity -- labeled with \textsc{o}).
%
%
%
%
Features can be defined over such edges, where each feature is assigned a weight.
Such features, together with their weights, can be used to score each possible path in the figure.
When dependency trees are provided, one can define features involving some local dependency structures.
For example, for the word ``{\em Womack}'', one possible feature that can be defined over edges covering this word can be of the form ``{\em Womack }$\leftarrow${\em won}'', indicating there is an incoming arc from the word ``{\em won}'' to the current word ``{\em Womack}''.
However, we note that such features largely ignore the global structured information as presented by the dependency trees.
Specifically, certain useful facts such as the word ``{\em Lee}'' is a child of the word ``{\em Womack}'' and at the same time a grandchild of the word ``{\em won}'' is not captured due to such a simple treatment of dependency structures.

\subsection{Dependency-Guided NER}


The main question we would like to ask is: how should we make good use of the structured information associated with the dependency trees to perform named entity recognition?
Since we are essentially interested in NER only, would there be some more global structured information in the dependency trees that can be used to guide the NER process?

From the earlier two examples shown in  Figure \ref{fig:twoexample} as well as the example shown in Figure \ref{fig:graphexample} we can observe that the named entities tend to be covered by a single or multiple adjacent dependency arcs.
This is not a surprising finding as for most named entities which convey certain semantically meaningful information, 
it is expected that there exist certain internal structures -- {\em i.e.}, dependencies amongst different words within the entities.
Words inside each named entity typically do not have dependencies with words outside the entities, except for certain words such as head words which typically have incoming arcs from outside words.

This finding motivates us to exploit such global structured information as presented by dependency trees for performing  NER.
Following the above observation, we first introduce the following notion:
\begin{definition}[Definition 1]
	\label{def:1}
	A {\em valid span} either consists of a single word, or is a word sequence that is covered by a  chain of (undirected) arcs where no arc is covered by another.
\end{definition}

For example, in Figure \ref{fig:graphexample}, the word sequence ``{\em Lee Ann Womack}'' is a valid span since there exists a single arc from ``{\em Womack}'' to ``{\em Lee}''. The sequence ``{\em Ann Womack won}'' is also a valid span due to a chain of undirected arcs -- one between ``{\em Ann}'' and ``{\em Womack}'' and the other between ``{\em Womack}'' and ``{\em won}''.
Similarly, the single word ``{\em Womack}'' is also a valid span given the above definition.

Based on the above definition, we can build a novel dependency-guided NER model based on the conventional semi-CRFs  by restricting the space of all possible combinations of entities to those that strictly contain only valid spans.
This leads to the following new partition function:
\begin{equation}
Z(\vec{x}) = \!\!\!\!\!\!\sum_{(i,j)\in\mathcal{S}_L(\vec{x})}\sum_{y'\in T} \sum_{y\in T}\exp(\vec{w}^{T}\vec{f}(\vec{x}, y', y, {i}, {j}) )
\end{equation}
where $\mathcal{S}(\mathbf{x})$ refers to the set of valid spans for a given input sentence $\mathbf{x}$, and $\mathcal{S}_L(\mathbf{x})$ refers to its subset that contains only those valid spans whose lengths are no longer than $L$.

We call the resulting model {\em dependency-guided model} (\textsc{dgm}).
Figure \ref{fig:graphexample} (middle) presents the graphical illustration of all possible paths (combination of entities) with $L=4$ that our model considers.
For example, since the word sequence ``{\em Ann Womack won Single}'' is not a valid span, the corresponding edges covering this word sequence is removed from the original lattice representation in Figure \ref{fig:graphexample} (top).
The edge covering the word sequence ``{\em of the Year}'' remains there as it is covered by the arc from ``{\em of}'' to ``{\em Year}'', thus a valid span.
As we can see, the amount of paths that we consider in the new model is substantially reduced.

\subsection{Time Complexity}
\begin{figure}[t!]
\small
	\begin{subfigure}[t]{0.51\columnwidth}
		\begin{tikzpicture}[node distance=3mm and 7mm, >=Stealth]
		\node[place,line width=1pt, minimum size=0.2cm] (p1) {};
		\node[place,line width=1pt, right=of p1, minimum size=0.2cm] (p2) {};
		\node[place,line width=1pt, right=of p2, minimum size=0.2cm] (p3) {};
		\node[place,line width=1pt, right=of p3, minimum size=0.2cm] (p4) {};
		\node[place,line width=1pt, right=of p4, minimum size=0.2cm] (p5) {};
		\draw [line width=1pt, -{Stealth[length=3.5mm, open]},->] (p5) to [out=120,in=50, looseness=0.7] node [above] {} (p1);
		\draw [line width=1pt, -{Stealth[length=3.5mm, open]},->] (p5) to [out=120,in=60, looseness=0.7] node [above] {} (p2);
		\draw [line width=1pt, -{Stealth[length=3.5mm, open]},->] (p5) to [out=120,in=60, looseness=0.7] node [above] {} (p3);
		\draw [line width=1pt, -{Stealth[length=3.5mm, open]},->] (p5) to [out=120,in=60, looseness=1.1] node [above] {} (p4);
		\end{tikzpicture}
		\caption{Best-case Scenario}
		\label{fig:bestcase} 
	\end{subfigure}
	\begin{subfigure}[t]{0.48\columnwidth}
		\begin{tikzpicture}[node distance=3mm and 7mm, >=Stealth]
		\node[place,line width=1pt, minimum size=0.2cm] (p1) {};
		\node[place,line width=1pt, right=of p1, minimum size=0.2cm] (p2) {};
		\node[place,line width=1pt, right=of p2, minimum size=0.2cm] (p3) {};
		\node[place,line width=1pt, right=of p3, minimum size=0.2cm] (p4) {};
		\node[place,line width=1pt, right=of p4, minimum size=0.2cm] (p5) {};
		\draw [line width=1pt, -{Stealth[length=3.5mm, open]},->] (p5) to [out=120,in=50, looseness=1] node [above] {} (p4);
		\draw [line width=1pt, -{Stealth[length=3.5mm, open]},->] (p4) to [out=120,in=50, looseness=1] node [above] {} (p3);
		\draw [line width=1pt, -{Stealth[length=3.5mm, open]},->] (p3) to [out=120,in=50, looseness=1] node [above] {} (p2);
		\draw [line width=1pt, -{Stealth[length=3.5mm, open]},->] (p2) to [out=120,in=50, looseness=1] node [above] {} (p1);
		\end{tikzpicture}
		\caption{Worst-case Scenario}
		\label{fig:worstcase} 
	\end{subfigure}
	\caption{The best-case and worst-case scenarios of  \textsc{dgm}.}
	\label{fig:scenarios_analysis}
\end{figure}
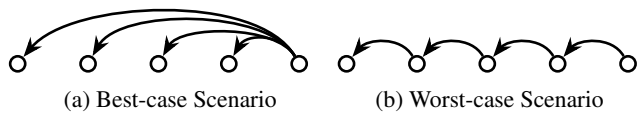

We can also analyze the time complexity required for this new model. We show example best-case and worst-case scenarios in Figure~\ref{fig:scenarios_analysis}.
For the former case there are $\mathcal{O} (n)$ valid spans. Thus the time complexity in the best case is $\mathcal{O} (n|T|^2)$, the same as the linear-chain CRFs model.
For the latter, there are $\mathcal{O} (nL)$ valid spans, leading to the time complexity $\mathcal{O} (nL|T|^2)$ in the worst case, the same as the semi-Markov CRFs model.

Furthermore, we have the following theorem for the average-case time complexity:
\begin{theorem}
The average-case time complexity of \textsc{dgm} is $\mathcal{O}(n|T|^2)$.
\end{theorem}
\begin{proof}
We provide two proofs to the above theorem. Below we show one proof and we move the second proof to the supplementary material (S.3.1).
	In this proof we assume the largest possible value for $L$, which is $n$ ({\em i.e.}, we do not limit the length of valid spans). Since the time complexity of our model depends directly on the total number of edges in the \textsc{dgm} graph, which is the same as the number of valid spans, we count the total number of valid spans over all possible structures by defining a bijection between each valid span and \textit{a tree} with $(n+1)$ nodes. Here we consider only one entity type since when we have $\left\lvert T\right\rvert$ types, the number of edges can be simply calculated by multiplying $\left\lvert T\right\rvert^2$ to the number that we will calculate here.
	
	Let $\mathcal{T}(n)$ be the set of all undirected trees with $n$ nodes, let $\tau\in\mathcal{T}(n)$ be an undirected tree with $n$ nodes, let $s_{\tau}^{u,v}$ be a valid span based on $\tau$ covering the words from position $u$ to $v$, and let $\mathcal{S}(\tau) = \{s_{\tau}^{u,v}\,|\,1\leq u\leq v\leq n\}$ be the set of valid spans based on $\tau$. Then the set of all valid spans over all possible undirected trees with $n$ nodes is $\mathcal{S}_{\textsc{dgm}}(n) =\cup_{\tau\in\mathcal{T}(n)}\mathcal{S}(\tau)$. Note that since the definition of valid spans uses undirected arcs of the dependency tree, we use undirected trees to count the number of valid spans.
	
	Now we define the bijection from $\mathcal{S}_{\textsc{dgm}}(n)$ to $\mathcal{T}(n+1)$ to show that the two sets have the same number of elements. By definition of valid spans, for each $s_{\tau}^{u,v} \in \mathcal{S}_{\textsc{dgm}}(n)$, there is exactly one associated chain of edges in the dependency tree $\tau$, which is the chain of edges starting at $u$ and ending at $v$, possibly visiting some intermediary nodes in between.
	
	Then, to define the bijection we map $s_{\tau}^{u,v}$ to a tree $\tau' \in \mathcal{T}(n+1)$ which is constructed from $\tau$ as follows: the edges for the first $n$ nodes coincide with $\tau$, except that the edges involved in the chain of edges are removed, and we make the nodes involved in this chain of edges to be connected to the $(n+1)$-th node instead. The process creates a well-defined mapping since the resulting graph is still connected and has $n$ edges over $(n+1)$ nodes, which means it is a tree with $(n+1)$ nodes.
	
	Conversely, from a tree $\tau' \in \mathcal{T}(n+1)$ we can recover the valid span and the undirected tree $\tau$ it is based on by removing the $(n+1)$-th node and the associated edges, then connecting the nodes that were previously connected to the $(n+1)$-th node in increasing order. The smallest and largest positions of the nodes define the starting and ending position of the span, and the resulting tree is the undirected tree $\tau$.
	
	For illustration, Figure \ref{fig:validspan} shows an example of a valid span $s_{\tau}^{1,4}$ where $\tau\in\mathcal{T}(4)$ and the nodes and edges involved in defining the valid span are highlighted in red. Figure \ref{fig:mappedtree} shows the associated tree $\tau'\in\mathcal{T}(5)$ where the removed edges are put in dashed lines and the new edges in bold.
	
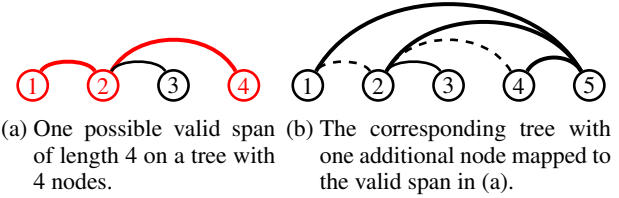
\begin{figure}[t!]
\small
	\begin{subfigure}[t]{0.45\columnwidth}
		\centering
		\begin{tikzpicture}[node distance=2mm and 5mm, >=Stealth]
		\node[place,line width=1pt, minimum size=0.4cm,red] (p1) {1};
		\node[place,line width=1pt, right=of p1, minimum size=0.4cm,red] (p2) {2};
		\node[place,line width=1pt, right=of p2, minimum size=0.4cm] (p3) {3};
		\node[place,line width=1pt, right=of p3, minimum size=0.4cm,red] (p4) {4};
		\draw [line width=1.5pt, -{Stealth[length=3.5mm, open]},-,red] (p4) to [out=120,in=60, looseness=1] node [above] {} (p2);
		\draw [line width=1pt, -{Stealth[length=3.5mm, open]},-] (p3) to [out=120,in=60, looseness=0.7] node [above] {} (p2);
		\draw [line width=1.5pt, -{Stealth[length=3.5mm, open]},-,red] (p2) to [out=120,in=60, looseness=0.7] node [above] {} (p1);
		\end{tikzpicture}
		\captionsetup{format=hang,margin=2pt}
		\caption{One possible valid span of length 4 on a tree with 4 nodes.}
		\label{fig:validspan} 
	\end{subfigure}%
	\begin{subfigure}[t]{0.53\columnwidth}
		\centering
		\begin{tikzpicture}[node distance=2mm and 5mm, >=Stealth]
		\node[place,line width=1pt, minimum size=0.4cm] (p1) {1};
		\node[place,line width=1pt, right=of p1, minimum size=0.4cm] (p2) {2};
		\node[place,line width=1pt, right=of p2, minimum size=0.4cm] (p3) {3};
		\node[place,line width=1pt, right=of p3, minimum size=0.4cm] (p4) {4};
		\node[place,line width=1pt, right=of p4, minimum size=0.4cm] (p5) {5};
		\draw [line width=1pt, -{Stealth[length=3.5mm, open]},dashed,-] (p4) to [out=120,in=60, looseness=1] node [above] {} (p2);
		\draw [line width=1pt, -{Stealth[length=3.5mm, open]},-] (p3) to [out=120,in=60, looseness=0.7] node [above] {} (p2);
		\draw [line width=1pt, -{Stealth[length=3.5mm, open]},dashed,-] (p2) to [out=120,in=60, looseness=0.7] node [above] {} (p1);
		\draw [line width=1.5pt, -{Stealth[length=3.5mm, open]},-] (p5) to [out=120,in=60, looseness=1] node [above] {} (p4);
		\draw [line width=1.5pt, -{Stealth[length=3.5mm, open]},-] (p5) to [out=120,in=60, looseness=1] node [above] {} (p2);
		\draw [line width=1.5pt, -{Stealth[length=3.5mm, open]},-] (p5) to [out=120,in=60, looseness=1] node [above] {} (p1);
		\end{tikzpicture}
		\captionsetup{format=hang,margin=2pt}
		\caption{The corresponding tree with one additional node mapped to the valid span in (a).}
		\label{fig:mappedtree} 
	\end{subfigure}
	\caption{An illustration of the bijection between a valid span $s_{\tau}^{u,v}$ and a tree $\tau' \in \mathcal{T}(n+1)$.}
	\label{fig:proof_illustration}
\end{figure}
	
	Therefore the mapping is bijective, and thus the number of total valid spans in $\mathcal{S}_{\textsc{dgm}}(n)$ is equal to the number of trees in $\mathcal{T}(n+1)$. It is a known result that the number of possible trees with $n$ nodes is $n^{n-2}$ \cite[Chapter 30]{Aigner2010}, which means the total number of valid spans is $(n+1)^{n-1}$. Thus, the average number of valid spans is:
	\begin{equation}
	\frac{\mathcal{S}_{\textsc{dgm}}(n)}{\left\lvert \mathcal{T}(n)\right\rvert} = \frac{(n+1)^{n-1}}{n^{n-2}} = n\left(1+\frac{1}{n}\right)^{n-1} \leq n\cdot \mathrm{\mathbf{e}}
	\end{equation}
which is linear in terms of $n$, just like the linear-chain CRFs. Here $\mathrm{\mathbf{e}}$ is the Euler number ($\approx 2.718$).
	
	This shows that the average-case time complexity of our model is $O(n|T|^2)$.\qed
\end{proof}
	
Now, this is a remarkable result, showing that the  complexity of our \textsc{dgm} model in its average case is still linear in the sentence length $n$ even if we set $L$ to its maximal value $n$. This is also true empirically, which we show in the supplementary material (S.3.2).
So, while our model retains the ability to capture non-Markovian features like semi-Markov CRFs,
it is more scalable and can be used to extract longer entities.

Besides the standard \textsc{dgm} model defined above, we also consider a variant, where we restrict the  chain (of arcs) to be of length 1 ({\em i.e.}, single arc) only. We call the resulting model \textsc{dgm-s}, where \textsc{s} stands for {\em single arc}.
Such a simplified model will lead to an even simpler lattice representation, resulting in even less running time.
However it may also lead to  lower performance as such a  variant might not be able to capture certain named entities.
In one of the later sections, we will conduct experiments on such a model and compare its performance with other models.

\section{Experimental Setup}
For experiments, we followed~\cite{finkel2009joint} and used the Broadcast News section of the OntoNotes dataset.
Instead of using its earlier 2.0 release, we used the final release --  release 5.0 of the dataset, which is available for download\footnote{https://catalog.ldc.upenn.edu/LDC2013T19/}.
The dataset includes the following 6 subsections: ABC, CNN, MNB, NBC, PRI and VOA. 
Moreover, the current OntoNotes 5.0 release also includes some English P2.5 data, which consists of corpus translated from Chinese and Arabic.\footnote{The OntoNotes 5.0 dataset also contains the earlier release from OntoNotes 1.0 to OntoNotes 4.0. However, we found the number of sentences of the Broadcast News in the current OntoNotes 5.0 dataset does not match the number reported in \cite{finkel2009joint,finkel2010hierarchical}, which was based on OntoNotes 2.0. Furthermore, they removed some instances which are inconsistent with their model. We thus cannot conduct experiments to compare against the results reported in their work.}
Following~\cite{finkel2009joint}, we split the first 75\% of the data for training and performed evaluations on the remaining 25\%.
We set $L=8$, which can cover all entities in our dataset, and developed the $L_2$ coefficient using cross-validation (see supplementary material S.1 for details). 
Following previous works on dependency parsing~\cite{chen2014fast}, we preprocessed the dataset using the Stanford CoreNLP\footnote{http://stanfordnlp.github.io/CoreNLP/} to convert the constituency trees to basic Stanford dependency~\cite{de2006generating} trees. 
In our NER experiments, in addition to using the given dependency structures converted from the constituency trees, we also trained a popular transition-based parser, MaltParser\footnote{http://maltparser.org/}~\cite{nivre2006maltparser}, using the training set and then used this parser to predict the dependency trees on the test set to be used in our model. 
%

\begin{table}[t!]
	\centering
	\scalebox{0.89}{
		\begin{tabular}{l|r|rrr}
			\multirow{2}{*}{} & \multirow{2}{*}{\# Sent.} & \multicolumn{3}{c}{\# Entities}\\
			\cline{3-5}
			&&\multicolumn{1}{c}{\textsc{all}}&\multicolumn{1}{c}{\textsc{dgm-s}}&\multicolumn{1}{c}{\textsc{dgm}}\\
			\hline
			Train                   & 9,996                        & 18,855                         & 17,584 (93.3\%)               & 18,803 (99.7\%)                 \\
			Test			& 3,339                          & 5,742                         & 5,309 (92.5\%)               & 5,720 (99.6\%)               \\
		\end{tabular}
		}
	\caption{Dataset statistics. }
	\label{tab:statistics}
\end{table}

\begin{table*}[ht!]
	\centering
	\begin{tabular}{cr|ccccccc|c}
		Dependency	& Model                  & ABC                               & CNN                               & MNB                               & NBC                               & P2.5                              & PRI                               & VOA  &Overall                           \\ \hline
		\multirow{4}{*}{Given}&Linear-chain CRFs & 70.2                              & 75.9                              & \textbf{75.7}                     & 65.9                              & 70.8                              & 83.2                              & 84.6                              & 77.8                              \\
		&Semi-Markov CRFs  & \textbf{71.9}                     & \textbf{78.2}                     & \textbf{74.7}                     & \textbf{69.4}                     & 73.5                              & \textbf{85.1}                     & 85.4                              & 79.6                              \\
		&\textsc{dgm-s}            & 71.4                              & 77.0                              & 73.4                              & 68.4                              & 72.8                              & \textbf{85.1}                     & 85.2                              & 79.0                              \\
		& \textsc{dgm}              & \textbf{72.3}                     & \textbf{78.6}                     & \textbf{76.3}                     & \textbf{69.7}                     & \textbf{75.5}                     & \textbf{85.5}                     & \textbf{86.8}                     & \textbf{80.5}                     \\\hline \hline
		\multirow{4}{*}{Predicted}& Linear-chain CRFs & 68.4          & 75.4          & 74.4          & 66.3          & 70.8          & 83.3          & 83.7          & 77.3          \\
		& Semi-Markov CRFs  & \textbf{71.6} & \textbf{78.0} & 73.5          & \textbf{71.5} & \textbf{73.7}          & \textbf{84.6} & \textbf{85.3} & \textbf{79.5} \\
		&\textsc{dgm-s}            & 70.6          & 76.4          & 73.4          & 68.7          & 71.3          & \textbf{83.9} & 84.4          & 78.2          \\
		&\textsc{dgm}              & \textbf{71.9} & \textbf{77.6} & \textbf{75.4} & \textbf{71.4} & \textbf{73.9} & \textbf{84.2} & \textbf{85.1} & \textbf{79.4}

	\end{tabular}
	\caption{NER results for all models, when given and predicted dependency trees are used and dependency features are used. Best values and the values which are not significantly different in 95\% confidence interval are put in bold.}
	\label{tab:ner_withdpfeatures}
\end{table*}

\begin{table*}[ht!]
	\centering
	\begin{tabular}{cr|ccccccc|c}
		Dependency	&	Model                  & ABC           & CNN           & MNB           & NBC           & P2.5          & PRI           & VOA           & Overall       \\ \hline
		\multirow{4}{*}{Given}&Linear-chain CRFs       & \multicolumn{1}{c}{66.5}          & \multicolumn{1}{c}{74.1}          & \multicolumn{1}{c}{74.9}          & \multicolumn{1}{c}{65.4}          & \multicolumn{1}{c}{70.8}          & \multicolumn{1}{c}{82.9}          & \multicolumn{1}{c}{82.3}          & \multicolumn{1}{|c}{76.3}          \\
		&Semi-Markov CRFs        & \multicolumn{1}{c}{\textbf{72.3}} & \multicolumn{1}{c}{76.6}          & \multicolumn{1}{c}{\textbf{75.0}} & \multicolumn{1}{c}{\textbf{69.3}} & \multicolumn{1}{c}{73.7}          & \multicolumn{1}{c}{84.1}          & \multicolumn{1}{c}{83.3}          & \multicolumn{1}{|c}{78.5}          \\
		&\textsc{dgm-s}                  & \multicolumn{1}{c}{69.4}          & \multicolumn{1}{c}{76.1}          & \multicolumn{1}{c}{73.4}          & \multicolumn{1}{c}{68.0}          & \multicolumn{1}{c}{72.5}          & \multicolumn{1}{c}{85.2}          & \multicolumn{1}{c}{85.1}          & \multicolumn{1}{|c}{78.6}          \\
		&\textsc{dgm}                    & \textbf{72.7} & \multicolumn{1}{c}{\textbf{77.2}} & \multicolumn{1}{c}{\textbf{75.8}} & \multicolumn{1}{c}{\textbf{68.5}} & \multicolumn{1}{c}{\textbf{76.8}} & \multicolumn{1}{c}{\textbf{86.2}} & \multicolumn{1}{c}{\textbf{85.5}} & \multicolumn{1}{|c}{\textbf{79.9}} \\ \hline \hline

		\multirow{4}{*}{Predicted}&Linear-chain CRFs       & 66.5          & 74.1          & 74.9          & 65.4          & 70.8          & 82.9          & 82.3           & 76.3          \\
		&Semi-Markov CRFs        & \textbf{72.3} & \textbf{76.6} & \textbf{75.0} & \textbf{69.3} & \textbf{73.7} & 84.1          & 83.3          & \textbf{78.5} \\
		&\textsc{dgm-s}                  & 69.1          & 75.6          & 73.8          & 67.2          & 72.0          & 84.5          & \textbf{84.2} & 78.0          \\
		&\textsc{dgm}                    & 71.3          & \textbf{76.2} & \textbf{75.9} & \textbf{68.8} & \textbf{74.6} & \textbf{85.1} & \textbf{84.3} & \textbf{78.8}
	\end{tabular}
	
	\caption{NER results for all models, when  given and predicted dependency trees are used but dependency features are not used. Best values and the values which are not significantly different in 95\% confidence interval are put in bold.}
	\label{tab:ner_withoutdpfeatures}
\end{table*}

We also looked at the SemEval-2010 Task 1 OntoNotes English corpus\footnote{https://catalog.ldc.upenn.edu/LDC2011T01/}, which contains sentences with both dependency and named entity information.
Although this dataset is a subset of the OntoNotes dataset, it comes with manually annotated dependency structures.
We conducted experiments on this dataset and reported the results in the supplementary material (S.2).
The results on this dataset are consistent with the findings reported in this paper.


\subsection{Features}
In order to make comparisons, we implemented a linear-chain CRFs model as well as a semi-Markov CRFs model to serve as baselines. 
The features used in this paper are basic features which are commonly used in linear-chain CRFs and semi-CRFs. 
For the linear-chain CRFs, we consider the current word/POS tag, the previous word/POS tag, the current word shape, the previous word shape, prefix/suffix of length up to 3, as well as transition features. 
For word shape features, we followed \cite{finkel2005exploring} to create them. 
For the semi-CRFs model, we have the following features for each segment: the word/POS tag/word shape before and after the segment, indexed words/POS tags/word shapes in current segment, surface form of the segment, segment length, segment prefix/suffix of length up to 3, the start/end word/tags of current segment, and the transition features. 

Inspired by \citeauthor{ling2012fine} \shortcite{ling2012fine}, we applied the following dependency features for all models: (1) current word in current position/segment and its head word (and its dependency label); (2) current POS tag in current position/segment and its head tag (and  dependency label). More details of features can be found in the supplementary material (S.4).

\subsection{Data Statistics}
There are in total 18 well-defined named entity types in the OntoNotes 5.0 dataset. 
Since majority of the entities are from the following three types: \textsc{per} ({\em person}), \textsc{org} ({\em organization}) and \textsc{gpe} ({\em geo-political entities}), following \cite{finkel2009joint}, we merge all the other entity types into one general type, \textsc{misc} ({\em miscellaneous}). 
Table \ref{tab:statistics} shows the statistics of total number of sentences and entities.


To show the relationships between the named entities and dependency structures, we present the number and percentage of entities that can be handled by our \textsc{dgm-s} model and \textsc{dgm} model respectively.
The entities that can be handled by \textsc{dgm-s} should be covered by a single arc as indicated in our model section. As for \textsc{dgm}, the entity spans should be {\em valid} as in definition 1. 
Overall, we can see that 93.3\% and 92.5\% of the entities can be handled by the \textsc{dgm-s} model in training set and test set, respectively. 
These two numbers for \textsc{dgm} are much higher -- 99.7\% and 99.6\%. 
With the predicted dependency structures in test set, 91.7\% of the entities can be handled by the \textsc{dgm-s} model, while for \textsc{dgm} it is 97.4\%.

We provide more detailed statistics for each of the 7 subsections in the supplementary material (S.1).
These numbers confirm that it is indeed the case that most named entities do form {\em valid spans}, even when using predicted dependency trees,
and that such global structured information of dependency trees can be exploited for NER.

%

\section{Results and Discussions}

\subsection{NER Performance}
Following previous work~\cite{finkel2009joint}, we report standard  F-score in this section (detailed results with precision and recall can be found in the supplementary material S.5). 
Table \ref{tab:ner_withdpfeatures} shows the results of all models when  dependency features are exploited.
Specifically, it shows results when the given and predicted dependency trees are considered, respectively.
Overall, the semi-Markov CRFs, \textsc{dgm-s} and \textsc{dgm} models perform better than the linear-chain CRFs model. 
Our \textsc{dgm} model obtains an overall F-score at 80.5\%  and outperforms the semi-CRFs model significantly ($p<0.001$ with bootstrap resampling~\cite{koehn2004statistical}).
For individual subsection, our \textsc{dgm} also performs the best.
On 2 out of the 7 subsections, our model's improvements over the baseline semi-CRFs model are significant ($p<0.001$ with bootstrap resampling).
For some other subsections like ABC, CNN, MNB, NBC and PRI, \textsc{dgm} has higher F-score than the semi-CRFs model but the improvements are not statistically significant.
The results show that comparing with  semi-CRFs, our \textsc{dgm} model, which has a lower average-case time complexity, still maintains a competitive performance. 
Such results confirm that the global structured information conveyed by dependency trees are useful and can be exploited by our \textsc{dgm} model.

The performance of \textsc{dgm-s} is worse than that of semi-CRFs and \textsc{dgm} in general since there are still 
many named entities that can not be handled by such a simplified model (see Table \ref{tab:statistics}).
This model has the same time complexity as the linear-chain CRFs, but performs better than linear-chain CRFs, largely due to the fact that certain structured information of dependency trees are exploited in \textsc{dgm-s}.
We note that such a simplified model obtains similar results as semi-CRFs for the two larger subsections, PRI and VOA.
This is largely due to the fact that a larger percentage of the entities in these two subsections can be handled by \textsc{dgm-s} (see supplementary material S.1 for more details).
It is noted that the performance of both semi-CRFs and \textsc{dgm} degrades when the predicted dependency trees are used instead of the given.
The drop in F-score for \textsc{dgm} is more significant as compared to the semi-CRFs.
Nevertheless, their results remain comparable.
Such experiments show the importance of considering high quality dependency structures for performing guided NER in our model.

To understand the usefulness of the global structured information of dependency trees better, we conducted further experiments by excluding dependency features from our models.
Such results are shown in Table \ref{tab:ner_withoutdpfeatures}.
Our \textsc{dgm} model consistently performs competitively with the semi-CRFs model. 
The only exception occurs when the ABC subsection is considered and the predicted dependency trees are used ($p$=0.067).
In general, we can see that when dependency features are excluded, the overall F-score for all models drop substantially.
However, we note that for semi-CRFs, the drop in F-score is 1.1\% for given dependency trees, and is 1.0\% for predicted trees,
whereas for \textsc{dgm}, the drops with given and predicted trees are both 0.6\%.
Overall, such results largely show that our proposed  model is able to effectively make use of the global structured information conveyed by dependency trees for NER. 

We have also conducted experiments on the widely-used NER dataset, CoNLL-2003, using the Stanford dependency parser\footnote{http://nlp.stanford.edu/software/nndep.shtml} to generate the dependency trees. 
Using the same feature set that we describe in this paper, our models do not achieve the state-of-the-art results on this dataset. However, they still perform comparably with the semi-CRFs model, while requiring much lesser running time.
Note that since our goal in this paper is to investigate the usefulness of incorporating the dependency structure information for NER, we did not attempt to tune the feature set to get the best result on a specific dataset.
Also it is worth remarking that we still obtain a relatively good performance for our \textsc{dgm} model although the dependency parser is not trained within the dataset itself and that a correct dependency structure information is crucial for the \textsc{dgm} model.

\begin{figure}[t!]
	\begin{subfigure}[t]{0.48\columnwidth}
		\centering
		\begin{tikzpicture}[node distance=3mm and 6mm, >=Stealth, 	place/.style={draw=none, inner sep=0pt}]
		\node[place,line width=1pt, minimum size=0.2cm] (13) {\small USS};
		\node[place,line width=1pt, right=of p1, minimum size=0.2cm, xshift=-2mm] (14) {\small Cole};
		\node[place,line width=1pt, right=of p2, minimum size=0.2cm, xshift=-2mm] (15) {\small Navy};
		\node[place,line width=1pt, right=of p3, minimum size=0.2cm, xshift=-3mm] (16) {\small destroyer};
		\node [place](at) [below=of 13,yshift=2.65mm]{\scriptsize NNP};
		\node [place](bt) [below=of 14,yshift=2.65mm] {\scriptsize NNP};
		\node [place](ct) [below=of 15,yshift=3.0mm] {\scriptsize NNP};
		\node [place](dt) [below=of 16,yshift=3.0mm]{\scriptsize NNP};
		\node [place](ae) [below=of at,yshift=2.45mm]{\scriptsize \textsc{b-misc}};
		\node [place](be) [below=of bt,yshift=2.5mm] {\scriptsize \textsc{i-misc}};
		\node [place](ce) [below=of ct,yshift=2.5mm] {\scriptsize \textsc{b-org}};
		\node [place](de) [below=of dt,yshift=2.5mm]{\scriptsize \textsc{o}};
		\draw [line width=1pt, -{Stealth[length=3.5mm, open]},->] (14) to [out=120,in=60, looseness=1.1] node [above] {} (13);
		\draw [line width=1pt, -{Stealth[length=3.5mm, open]},->] (16) to [out=120,in=60, looseness=0.9] node [above] {} (14);
		\draw [line width=1pt, -{Stealth[length=3.5mm, open]},->] (16) to [out=120,in=60, looseness=1.1] node [above] {} (15);
		\end{tikzpicture}
		\caption{Given dependency tree}
		\label{fig:error_given} 
	\end{subfigure}
	\begin{subfigure}[t]{0.48\columnwidth}
		\centering
		\begin{tikzpicture}[node distance=3mm and 6mm, >=Stealth, 	place/.style={draw=none, inner sep=0pt}]
		\node[place,line width=1pt, minimum size=0.2cm, xshift=7mm] (13) {\small USS};
		\node[place,line width=1pt, right=of p1, minimum size=0.2cm, xshift=5mm] (14) {\small Cole};
		\node[place,line width=1pt, right=of p2, minimum size=0.2cm, xshift=5mm] (15) {\small Navy};
		\node[place,line width=1pt, right=of p3, minimum size=0.2cm, xshift=5mm] (16) {\small destroyer};
		\node [place](at) [below=of 13,yshift=2.65mm]{\scriptsize NNP};
		\node [place](bt) [below=of 14,yshift=2.65mm] {\scriptsize NNP};
		\node [place](ct) [below=of 15,yshift=3.0mm] {\scriptsize NNP};
		\node [place](dt) [below=of 16,yshift=3.0mm]{\scriptsize NNP};
		\node [place](ae) [below=of at,yshift=2.45mm]{\scriptsize \textsc{b-org}};
		\node [place](be) [below=of bt,yshift=2.5mm] {\scriptsize \textsc{i-org}};
		\node [place](ce) [below=of ct,yshift=2.5mm] {\scriptsize \textsc{i-org}};
		\node [place](de) [below=of dt,yshift=2.5mm]{\scriptsize \textsc{o}};
		\draw [line width=1pt, -{Stealth[length=3.5mm, open]},->] (15) to [out=120,in=60, looseness=0.9] node [above] {} (13);
		\draw [line width=1pt, -{Stealth[length=3.5mm, open]},->] (15) to [out=120,in=60, looseness=1.1] node [above] {} (14);
		\draw [line width=1pt, -{Stealth[length=3.5mm, open]},->] (16) to [out=120,in=60, looseness=1.1] node [above] {} (15);
		\end{tikzpicture}
		\caption{Predicted dependency tree}
		\label{fig:error_predicted} 
	\end{subfigure}
	\caption{The effect of different dependency parses on the output of \textsc{dgm}. These are taken out from a part of a sentence. The NER result in (a) is correct, while (b) is not.}
	\label{fig:error_analysis}
\end{figure}
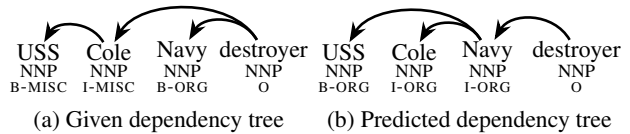
\subsection{Error Analysis}
We provide a further analysis of how the dependency parsing performance affects NER based on Table \ref{tab:ner_withoutdpfeatures}. 
Because our model uses the dependency structure information directly instead of using them as features, we can analyze the influence of dependency structures on NER more clearly.
Specifically, we focus on how the dependency parsing results affect the prediction of our \textsc{dgm} model. 

A typical error made by our model taken from the results is shown in Figure \ref{fig:error_analysis} where the dependency tree in Figure \ref{fig:error_given} is given by the conversion of constituent parse tree and the other one in Figure \ref{fig:error_predicted} is predicted from the MaltParser. 
The NER result on the left is correct while the one on the right is incorrect. 
Based on the predicted dependency structure in Figure \ref{fig:error_predicted}, there is no way to predict an entity type for ``{\em USS Cole}'' since this is not a valid span in \textsc{dgm} model. 
Furthermore, \textsc{dgm} can actually recognize ``{\em Navy}'' as an \textsc{org} entity even though the predicted dependency is incorrect. 
But the model considers ``{\em USS Cole Navy}'' as an entity due to the interference of other entity features ({\em e.g.}, NNP tag and Capitalized pattern) that ``{\em USS Cole}'' has. 
While with the given dependency tree, \textsc{dgm} considers ``{\em USS Cole}'' as a valid span and correctly recognizes it as a \textsc{misc} entity.

\subsection{Speed Analysis}
From Figure \ref{fig:graphexample} we can see that the running time required for each model depends on the number of edges that the model considers.
We thus empirically calculated the average number of edges per word each model considers based on our training data.
We found that the average number of edges involved in each token is 132 for the semi-CRFs model,
while this number becomes 40 and 61 for \textsc{dgm-s} and \textsc{dgm} respectively.
A lower number of edges indicates less possible structures to consider, and a reduced running time.
See more detailed information  in the supplementary material (S.3.2).

\begin{figure}[t!]
	\centering
	\scalebox{1}{
		\includegraphics[width=3.2in]{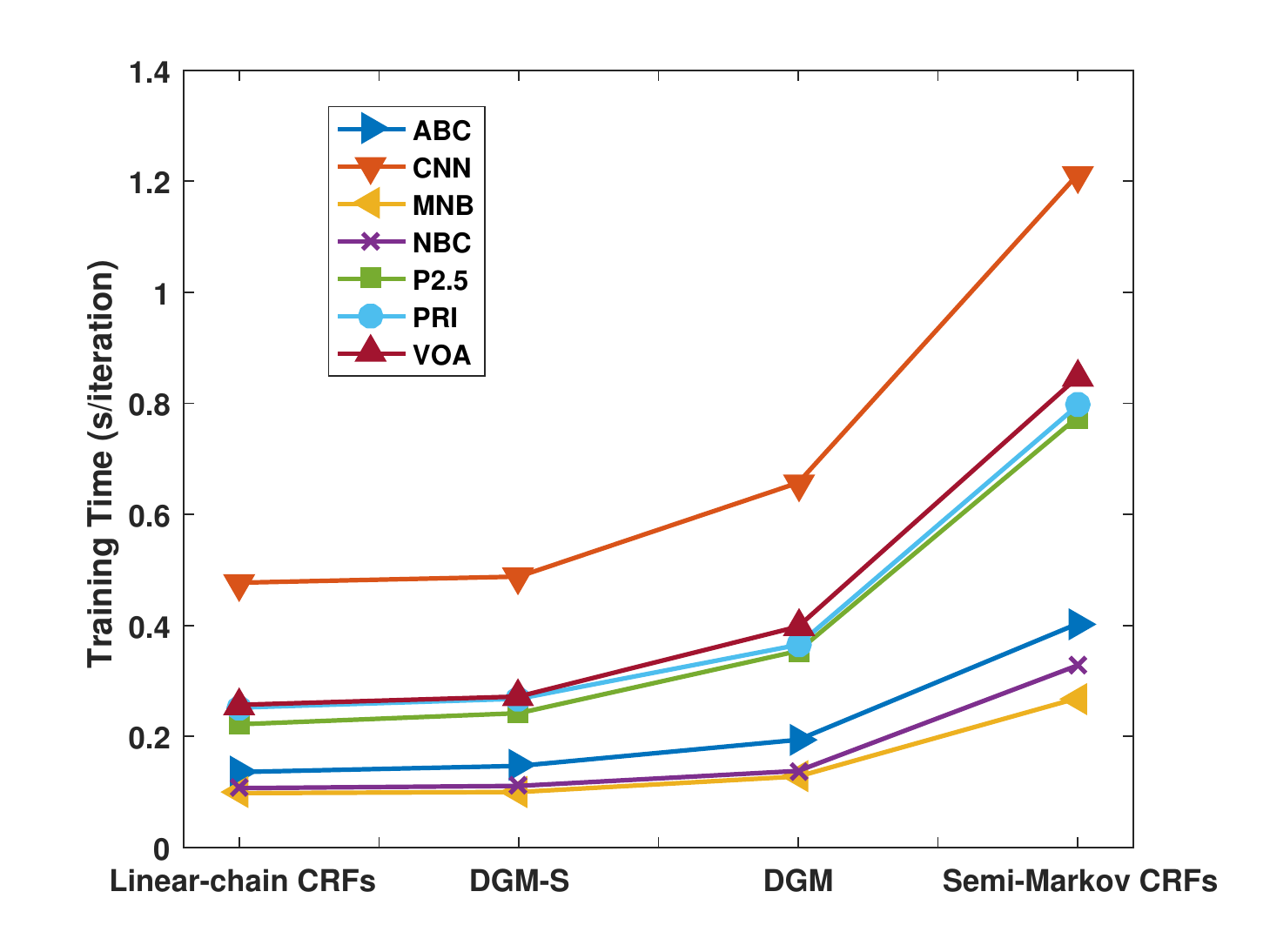}
	}
	\caption{Training time per iteration of all the models.}
	\label{fig:time}
\end{figure}

The results on training time per iteration (inference time) for all 7 subsections are shown in Figure \ref{fig:time}. 
From the figure we can see that the linear-chain CRFs model empirically runs the fastest.
The simple \textsc{dgm-s} model performs comparably with linear-chain CRFs.
The semi-CRFs model requires substantially longer time due to the additional factor $L$ (set to 8 in our experiments) in its time complexity.
In contrast, our model \textsc{dgm} requires only 47\% of the time needed for semi-CRFs for each training iteration, and requires 37\% more time than the \textsc{dgm-s} model.


\section{Conclusion}

In this paper, we proposed a novel efficient dependency-guided model for named entity recognition.
Motivated by the fact that named entities are typically covered by dependency arcs, presenting internal structures, we built a model that is able to explicitly exploit global structured information conveyed by dependency trees for NER.
We showed that the model theoretically is better than the semi-Markov CRFs model in terms of time complexity.
Experiments show that our model performs competitively with the semi-Markov CRFs model, even though it requires substantially less running time.
Future directions include further investigations on the structural relations between dependency trees and named entities,
and working towards building integrated models that perform joint prediction of both structures.
%

We make our code and system available for download at \texttt{http://statnlp.org/research/ie/}.

\section{Acknowledgments}
We would like to thank the anonymous reviewers for their constructive and helpful comments.
This work is supported by MOE Tier 1 grant SUTDT12015008.
We thank Razvan Bunescu for pointing out an error in Equations 3 and 5 in an earlier version of this paper.

\bibliography{dgner-arxiv}
\bibliographystyle{aaai}

\end{document}


\maketitle

\section{Detailed Data Statistics and Parameter Tuning}
Table \ref{tab:statistics} shows the detailed statistics of all subsections. 
Overall, over 99.6\% entities are representable in \textsc{dgm} model and around 91\% to 94\% entities are representable \textsc{dgm-s} model. 
\begin{table}[h]
	\small
	\centering
	\begin{tabular}{l||r|rrr||r|rrr}
		& \multicolumn{4}{|c|}{Train} & \multicolumn{4}{|c}{Test} \\\cline{2-9} 
		& \multirow{2}{*}{\# Sent.} & \multicolumn{3}{c|}{\# Entity} & \multirow{2}{*}{\# Sent.} & \multicolumn{3}{c}{\# Entity}\\
		\cline{3-5}\cline{7-9}
		&&\multicolumn{1}{c}{\textsc{all}}&\multicolumn{1}{c}{\textsc{dgm-s}}&\multicolumn{1}{c|}{\textsc{dgm}}  &&\multicolumn{1}{c}{\textsc{all}}&\multicolumn{1}{c}{\textsc{dgm-s}}&\multicolumn{1}{c}{\textsc{dgm}}\\
		\hline
		ABC                   & 873                          & 1,365                         & 1,281 (93.9\%)               & 1,360 (99.6\%)                	& 292                         & 444                           & 415 (93.5\%)               & 444 (100.0\%)                \\
		CNN                   & 4,318                        & 6,627                        & 6,113 (92.2\%)              & 6,609 (99.7\%)                & 1,440                        & 1,620                         & 1,474 (91.0\%)               & 1,613 (99.6\%)                \\
		MNB                   & 477                          & 690                           & 653 (94.6\%)               & 689 (99.9\%)                	& 160                          & 177                           & 162 (91.5\%)                & 176 (99.4\%)                \\
		NBC                   & 480                         & 922                           & 868 (94.1\%)               & 918 (99.6\%)                 & 161                          & 343                           & 312 (91.0\%)               & 340 (99.1\%)                \\
		P2.5                   & 890                          & 1,995                          & 1,827 (91.6\%)               & 1,988 (99.7\%)                 	& 298                          & 672                           & 616 (91.7\%)               & 669 (99.6\%)                \\
		PRI                   & 1,536                        & 3,096                         & 2,916 (94.2\%)               & 3,090 (99.8\%)                 	& 513                          & 992                           & 927 (93.5\%)               & 990 (99.8\%)                \\
		VOA                   & 1,422                        & 4,160                         & 3,926 (94.4\%)               & 4,149 (99.7\%)                 	& 475                          & 1,494                         & 1,403 (93.9\%)               & 1,488 (99.6\%)               \\	
		Total                   & 9,996                        & 18,855                         & 17,584 (93.3\%)               & 18,803 (99.7\%)                 & 3,339                          & 5,742                         & 5,309 (92.5\%)               & 5,720 (99.6\%)               \\
		\hline
	\end{tabular}
	\caption{Dataset Statistics. The number of sentences and entities in the Broadcast News corpus of OntoNotes 5.0 dataset. }
	\label{tab:statistics}
\end{table}
We tuned the $L_{2}$ regularization parameter using 10-fold cross-validation for all the models. 
Specifically for each model, we performed cross validation on the largest subsection, CNN, and obtained the best parameter with highest average F-score. 
We then used this parameter for other subsections as well. 
The values of $L_{2}$ regularization parameter we evaluated is $[0.0001, 0.001, 0.01, 0.1, 1]$. Specifically, we have four models and each of them is associated with two variants with or without dependency features. 
We obtained the best regularization parameter 0.1 for all the models except the \textsc{dgm-s} without dependency features, whose best regularization parameter is 1. 

\section{Results on SemEval-2010 Task 1 Dataset}
	
The dataset in SemEval-2010 Task 1 is a subset of OntoNotes English corpus. The dependency and named entities information are annotated in this dataset. 
There are a total of 3,648 sentences with 4,953 entities in the training set, 741 sentences with 1,165 entities in the development set, and 1,141 sentence with 1,697 entities in the test set. 
In this dataset, we found that overall 92.7\% of entities are representable in \textsc{dgm-s} and 96.6\% of entities are representable in \textsc{dgm}. 

Table \ref{tab:semres} and Table \ref{tab:semnodepres} show the NER performance of all models on SemEval 2010 Task 1 dataset. In Table \ref{tab:semres}, the semi-CRF model with gold dependency features has a higher F-score while it is not significantly better than our \textsc{dgm} ($p=0.44$). 
However, \textsc{dgm} performs worse when the predicted dependency is used as features since the percentages of entities representable in \textsc{dgm} is not as high as 99\% in the other dataset, and also the predicted dependency information affects our \textsc{dgm} model structures. Furthermore, if we do not use the dependency features, both \textsc{dgm} and semi-CRFs perform similarly, with p-values of $p=0.23$ and $p=0.33$ when using gold and predicted dependency structures for \textsc{dgm}, respectively. The Semi-CRF model achieves 74.5\% F-score while \textsc{dgm} (using gold dependency structures) achieves 74.8\% F-score with a much faster training time per iteration (inference time). The inference time of semi-CRFs on this dataset is 2.25 times more than the inference time of \textsc{dgm}. 
\begin{table*}[t!]
	\centering
	\resizebox{\columnwidth}{!}{
	\begin{tabular}{l|c|lll|lll|lll|lll}
		& \multirow{2}{*}{Dependency}& \multicolumn{3}{c|}{Linear-chain CRFs}                                   & \multicolumn{3}{c|}{Semi-Markov CRFs}                                   & \multicolumn{3}{c|}{\textsc{dgm-s}}                                              & \multicolumn{3}{c}{\textsc{dgm}}                                              \\
		& & \multicolumn{1}{c}{\em P} & \multicolumn{1}{c}{\em R} & \multicolumn{1}{c|}{\em F} & \multicolumn{1}{c}{\em P} & \multicolumn{1}{c}{\em R} & \multicolumn{1}{c|}{\em F} & \multicolumn{1}{c}{\em P} & \multicolumn{1}{c}{\em R} & \multicolumn{1}{c|}{\em F} & \multicolumn{1}{c}{\em P} & \multicolumn{1}{c}{\em R} & \multicolumn{1}{c}{\em F} \\ \hline
		\multirow{2}{*}{SemEval 2010}&Gold & 	75.8&	72.2&	73.9&77.3&73.8&\textbf{75.5}	&	76.1&	72.4&	74.2&	77.0&	73.0&	\textbf{75.0} \\
		&Predicted & 	75.2&	71.1&	73.1&77.2&73.2&\textbf{75.1}	&	76.0&	72.2&	74.1&	76.4&	71.8&	74.1          
	\end{tabular}
}
	\caption{Named Entity Recognition Results on the SemEval 2010 Task 1 dataset. All the models in this table use the dependency information as features.}
	\label{tab:semres}
\end{table*}

\begin{table*}[t!]
	\centering
	\resizebox{\columnwidth}{!}{
		\begin{tabular}{l|c|lll|lll|lll|lll}
			& \multirow{2}{*}{Dependency}& \multicolumn{3}{c|}{Linear-chain CRFs}                                   & \multicolumn{3}{c|}{Semi-Markov CRFs}                                   & \multicolumn{3}{c|}{\textsc{dgm-s}}                                              & \multicolumn{3}{c}{\textsc{dgm}}                                              \\
			& & \multicolumn{1}{c}{\em P} & \multicolumn{1}{c}{\em R} & \multicolumn{1}{c|}{\em F} & \multicolumn{1}{c}{\em P} & \multicolumn{1}{c}{\em R} & \multicolumn{1}{c|}{\em F} & \multicolumn{1}{c}{\em P} & \multicolumn{1}{c}{\em R} & \multicolumn{1}{c|}{\em F} & \multicolumn{1}{c}{\em P} & \multicolumn{1}{c}{\em R} & \multicolumn{1}{c}{\em F} \\ \hline
			\multirow{2}{*}{SemEval 2010}&Gold & 	\multirow{2}{*}{76.1}&	\multirow{2}{*}{71.2}&		\multirow{2}{*}{73.6}&\multirow{2}{*}{77.2}&\multirow{2}{*}{72.1}&\multirow{2}{*}{\textbf{74.5}	}&	77.1&	71.6&	74.3& 77.7  &	72.1& \textbf{74.8} \\ 
			&Predicted & 	&	&	&&&	&	77.0&	71.2&	74.0&	77.3&	71.5&	\textbf{74.3}     
		\end{tabular}
	}
	\caption{Named Entity Recognition Results on the SemEval 2010 Task 1 dataset without dependency features. Note that \textsc{dgm-s} and \textsc{dgm} still utilize the dependency information to build the models.}
	\label{tab:semnodepres}
\end{table*}

\section{Number of Edges}
The number of edges in a graphical model is proportional to the time complexity of training the model, and so in the interest of calculating the time complexity of the models, in this section we show theoretically that on average case, the number of edges in our \textsc{dgm} model is linear in terms of $n$, the number of words in a sentence. Then for all models we show empirically the relationship between the number of edges and the sentence length $n$.

\subsection{Theoretical Analysis}
In this section we show how to calculate the average number of edges in the \textsc{dgm} model for a sentence of length $n$. Note that the number of edges in the model is by definition equal to the number of valid spans, as an edge is added in the model connecting the beginning of the first word and the end of the last word of each valid span. In the first analysis, we will count the valid spans that cover more than one words separately from the valid spans covering only one word. Since the valid spans covering one word are always present, the count is known, there are $n$ of them for each graph in the model. So in the first method we count only the valid spans covering more than one words.

We first calculate the total number of edges in the \textsc{dgm} model over all possible undirected trees, then the average is just this number divided by the number of undirected trees. 
Although dependency trees are directed trees, calculating the average on undirected trees is sufficient as for each undirected tree there is exactly $n$ possible dependency trees, one for each selection of node as the root of the tree, and so the average number of edges will be the same. Note that we consider both projective and non-projective trees here. Later the empirical count will provide an evidence that the average number of edges calculated here still holds even in a dataset where most of the trees are projective.

More formally, if the vertex set for the graph $G_{\tau}$ in the \textsc{dgm} model with the undirected dependency tree $\tau$ is $V(G_{\tau}) = \{1,2,\ldots,n\}$ and the edge set of the tree is $E(\tau)$, then the edge set of the graph in the model is $E_{\textsc{dgm}}(G_{\tau}) = \{(u_1,u_{k+1})\ |\ \exists\ u_1<u_2<\ldots<u_{k+1}\ \text{s.t.}\ (u_i,u_{i+1})\in E(\tau),\ \forall i=1,2,\ldots,k,\ k\geq 1\}$. We present two alternative methods to count the average, the first using algebra, and the second using bijection. Although the first method is more complicated, along the process it also calculates the relationship between the average number of edges and $L$, the maximum valid span length. In this supplementary material, we present only the first method. The proof by bijection is already covered in the main paper \cite{Jie2017}.

\subsubsection{Proof by Algebra}

To count the number of edges over all possible trees, we consider each pair of nodes $u$ and $v$ in the graph, where $u<v$. Let $f_n(u,v)$ be the number of times there is an edge in the \textsc{dgm} model, over all possible trees. More formally, $f_n(u,v) = \sum_{\tau\in\mathcal{T}(n)} I\left[(u,v)\in E_{\textsc{dgm}}(G_{\tau})\right]$, where $I[F]$ is the indicator function, which is $1$ if the expression $F$ is true, $0$ otherwise, and $\mathcal{T}(n)$ is the set of all possible trees with $n$ nodes. Also, let $F(n, L)$ as the total number of edges in \textsc{dgm} model where the maximum valid span length is $L$. We are looking for $\bar{F}(n,n)$, the average number of edges in \textsc{dgm} in all possible trees when $L$ is not restricted.

First, to calculate $f_n(u,v)$, we consider the number of intermediate nodes between $u$ and $v$. Note that, by definition of the $E_{\textsc{dgm}}(G_{\tau})$, the intermediate nodes must appear between $u$ and $v$, and for any set of intermediate nodes there is exactly one set of edges that leads to $(u,v)$ being a valid span, that is when there is no edge covered by another. If $k$ is the number of intermediate nodes, then there are $\binom{v-u-1}{k}$ ways to choose the path from $u$ to $v$, and we can multiply this by the number of trees that contain each path to get $f_n(u,v)$. Note that for a path with $k$ intermediate nodes, there are $k+2$ nodes, including $u$ and $v$, among the $n$ nodes which are already connected. If we remove these edges from the tree that contains the path, we will be left with a forest with $k+2$ components, as each of the $k+2$ nodes will be disconnected from each other, due to the property of trees that there is a unique path between any two nodes. So the number of trees that contain a path with $k+2$ nodes is equal to the number of forest with $n$ nodes and $k+2$ components, which is well-established as $(k+2)n^{n-k-3}$ (see, for example \cite[Chapter 30]{Aigner2010}).

So we have:
\begin{eqnarray}
f_n(u,v) &=& \sum_{k=0}^{v-u-1}\binom{v-u-1}{k}(k+2)n^{n-k-3}\nonumber\\
&=& \sum_{k=0}^{v-u-1}\binom{v-u-1}{v-u-1-k}(v-u-1-k+2)n^{n-(v-u-1-k)-3}\nonumber\\
&=& n^{n-v+u-2}\sum_{k=0}^{v-u-1}\binom{v-u-1}{k}(v-u+1-k)n^k\nonumber\\
&=& n^{n-v+u-2}\left[(v-u+1)\sum_{k=0}^{v-u-1}\binom{v-u-1}{k}n^k - \sum_{k=0}^{v-u-1}\binom{v-u-1}{k}kn^k\right]\\
&=& n^{n-3-(v-u-1)}\left[(v-u+1)(n+1)^{v-u-1} - n(v-u-1)(n+1)^{v-u-2}\right]\\
&=& n^{n-3}\left[(v-u+1)\left(1+\frac{1}{n}\right)^{v-u-1} - (v-u-1)\left(1+\frac{1}{n}\right)^{v-u-2}\right]\nonumber\\
&=& n^{n-3}\left[\left(v-u+1 - \frac{n(v-u-1)}{n+1}\right)\left(1+\frac{1}{n}\right)^{v-u-1}\right]\nonumber\\
&=& n^{n-3}\left(\frac{2n+1+v-u}{n+1}\right)\left(1+\frac{1}{n}\right)^{v-u-1}\nonumber
\end{eqnarray}

To get from (1) to (2) we used the expansion of binomial formula and its derivative, as follows:
\begin{align*}
\qquad\qquad\qquad\qquad\qquad\qquad&&(x+1)^n &= \sum_{k=0}^n\binom{n}{k}x^k&&\nonumber\\
\qquad\qquad\qquad\qquad\qquad\qquad&&n(x+1)^{n-1} &= \sum_{k=0}^n\binom{n}{k}kx^{k-1}&&\text{(take derivative both sides)}\nonumber\\
\qquad\qquad\qquad\qquad\qquad\qquad&&nx(x+1)^{n-1} &= \sum_{k=0}^n\binom{n}{k}kx^k&&\text{(multiply by $x$)}\nonumber
\end{align*}
\noindent with $n$ and $x$ substituted for $v-u-1$ and $n$ respectively to get from (1) to (2). Note that since the formula depends on $u$ and $v$ only from the difference $v-u$, let us define $f_n(k) = f_n(u, u+k)$.

Now, the number of total edges is the sum of $f_n(u,v)$ over all possible values for $u$ and $v$, yielding our final formula for $F(n, L)$, the total number of edges when the maximum valid span is of length $L$:
\begin{eqnarray}
F(n,L) \!\!&=& \sum_{u=1}^{n-1}\sum_{v=u+1}^{u+L-1} f_n(u,v)\nonumber\\
&=& \sum_{k=1}^{L-1}\sum_{u=1}^{n-k} f_n(k)\nonumber\\
&=& \sum_{k=1}^{L-1}(n-k)n^{n-3}\left(\frac{2n+1+k}{n+1}\right)\left(1+\frac{1}{n}\right)^{k-1}\nonumber\\
&=& \frac{n^{n-3}}{n+1}\sum_{k=1}^{L-1}(n-1-(k-1))(2n+2+(k-1))\left(1+\frac{1}{n}\right)^{k-1}\nonumber\\
&=& \frac{n^{n-3}}{n+1}\sum_{k=0}^{L-2}(n-1-k)(2n+2+k)\left(1+\frac{1}{n}\right)^k\nonumber\\
&=& \frac{n^{n-3}}{n+1}\left[(n-1)(2n+2)\sum_{k=0}^{L-2}\left(1+\frac{1}{n}\right)^k - (n+3)\sum_{k=0}^{L-2}k\left(1+\frac{1}{n}\right)^k - \sum_{k=0}^{L-2}k^2\left(1+\frac{1}{n}\right)^k\right]\ \ \ \ \\
&=& \frac{n^{n-3}}{n+1}\left[(n-1)(2n+2)n\left(\left(1+\frac{1}{n}\right)^{L-1}-1\right)\right.\nonumber\\
&& -(n+3)\left(n^2(L-2)\left(1+\frac{1}{n}\right)^{L}-n^2(L-1)\left(1+\frac{1}{n}\right)^{L-1}+n^2\left(1+\frac{1}{n}\right)\right)\nonumber\\
&& \left.-(n(L-2)^2-2n^2(L-2)+2n^3+n^2)\left(1+\frac{1}{n}\right)^{L-1}+n^3\left(1+\frac{1}{n}\right)^2+n^3\left(1+\frac{1}{n}\right)\right]\\
&=& \frac{n^{n-3}}{n+1}\left[n(n^2+L(n-L+1))\left(1+\frac{1}{n}\right)^{L-1}-n^2(n+1)\right]\nonumber\\
&=& \frac{n^{n-2}}{n+1}\left[(n^2+L(n-L+1))\left(1+\frac{1}{n}\right)^{L-1}-n(n+1)\right]\nonumber
\end{eqnarray}
\noindent which, when $L = n$, that is when we do not restrict the valid span length, we have:
\begin{eqnarray}
F(n, n) &=& n^{n-1}\left[\left(1+\frac{1}{n}\right)^{n-1}-1\right] = (n+1)^{n-1} - n^{n-1}\nonumber
\end{eqnarray}

We used geometric sums formula and its derivatives (equations (5), (6), and (7)) to get from (3) to (4):
\begin{eqnarray}
\frac{x^{n+1}-1}{x-1} \!\!&=&\!\! \sum_{k=0}^n x^k\\
\text{(take derivative, multiply by }x\text{)}\qquad\frac{(n+1)x^{n+1}(x-1) - x(x^{n+1}-1)}{(x-1)^2} \!\!&=&\!\! \sum_{k=0}^n kx^k\nonumber\\
\text{(then rearrange. Repeat these two steps to get to (7))}\qquad\frac{nx^{n+2}-(n+1)x^{n+1}+x}{(x-1)^2} \!\!&=&\!\! \sum_{k=0}^n kx^k\\
x\frac{(n(n+2)x^{n+1}-(n+1)^2x^n+1)(x-1)^2 - 2(nx^{n+2}-(n+1)x^{n+1}+x)(x-1)}{(x-1)^4} \!\!&=&\!\! \sum_{k=0}^n k^2x^k\nonumber\\
x\frac{(n(x-1)(n(x-1)-2)+x+1)x^n-x-1}{(x-1)^3} \!\!&=&\!\! \sum_{k=0}^n k^2x^k\quad
\end{eqnarray}

Finally, the average of number of edges is $F(n, n)$, plus the number of valid spans covering one word, which is $n\cdot{}n^{n-2}$, divided by total number of trees, which is $T(n,1) = n^{n-2}$, which yields the average:

\begin{eqnarray}
\bar{F}(n) &=& \frac{((n+1)^{n-1}-n^{n-1}) + n^{n-1}}{n^{n-2}}\nonumber\\
&=& n\left(1+\frac{1}{n}\right)^{n-1}\nonumber\\
&\leq& \mathbf{e}n\nonumber
\end{eqnarray}
\noindent where $\mathbf{e}$ is the Euler's number, which is approximately 2.718.

So we can see that on average, the number of edges present in the \textsc{dgm} model for one entity type is linear in terms of $n$, the number of words in the sentence. When there are $\left\lvert T\right\rvert$ types, the number of edges will be multiplied by $\left\lvert T\right\rvert^2$ since for each edge there are $\left\lvert T\right\rvert^2$ possible type combination for the start and end of the span. {\bf Therefore the time complexity of the training process of \textsc{dgm} model when there are $\left\lvert T\right\rvert$ types is $\mathcal{O}(n\left\lvert T\right\rvert^2)$.}

\subsection{Empirical Count}
\begin{figure}[h]
	\centering
	\includegraphics[width=3.2in]{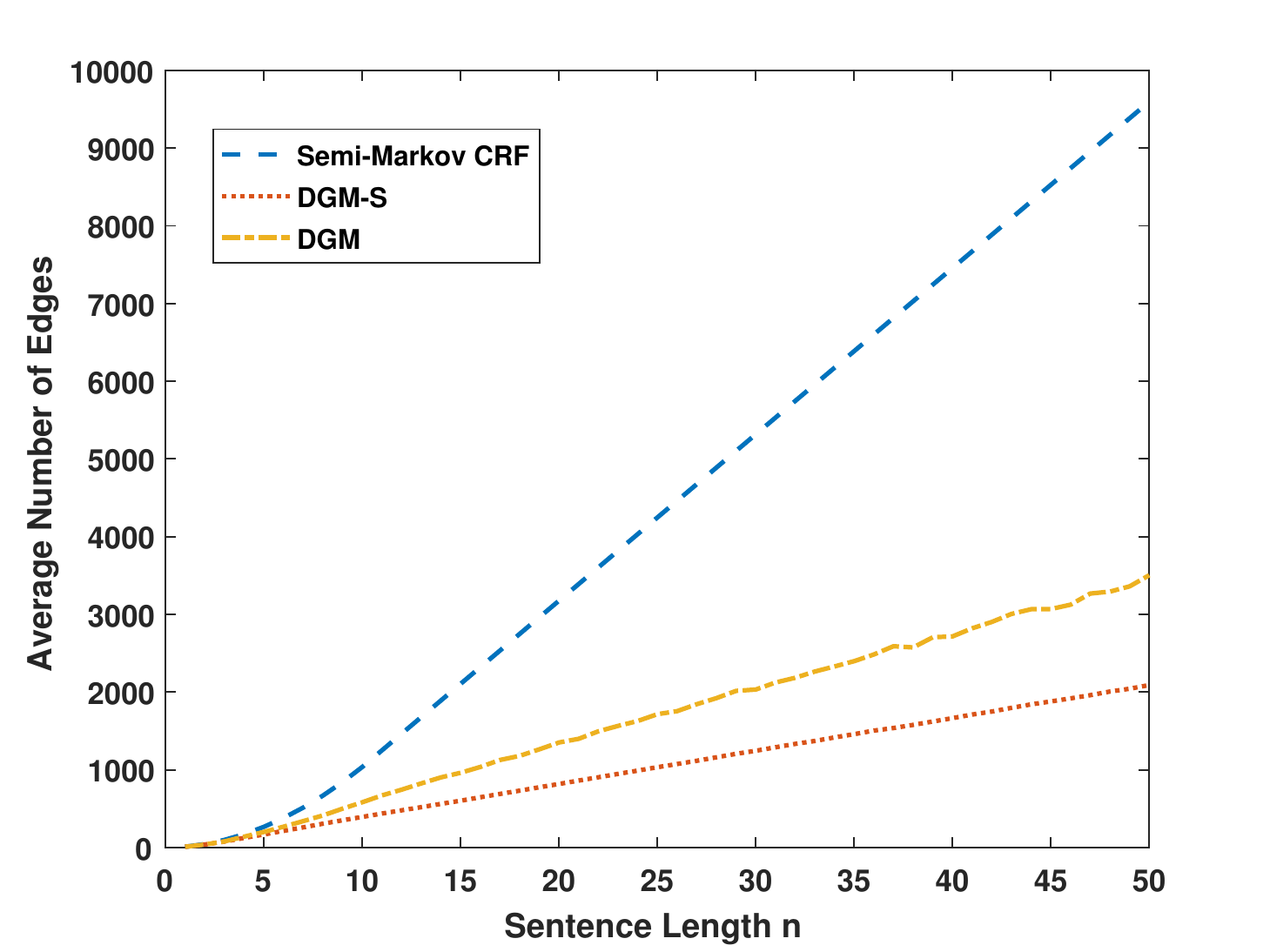}
	\caption{The average number of edges over all sentences in all datasets with respect to sentence length $n$. For semi-CRF, we set $L=8$. Also note that in the dataset we have $\left\lvert T\right\rvert=5$ (\textsc{per}, \textsc{org}, \textsc{gpe}, \textsc{misc}, and special label \textsc{o} denoting non-entities)}.
	\label{fig:ncomplexity}
\end{figure}

In this subsection, we calculate empirically the relationship between the number of edges present per sentence in each model and $n$, the sentence length, to provide an evidence for the theoretical analysis presented in previous subsection in a dataset where most of the trees are projective. We compute this by averaging the number of edges per sentence in all 7 subsections of the dataset. Figure \ref{fig:ncomplexity} shows the result. We can see that the number of edges in semi-CRFs model is much more than that of \textsc{dgm} and \textsc{dgm-s}. All three models have a linear complexity in terms of $n$. However, note that the semi-CRFs model comes with a scaling factor $L$, while our models do not. For the purpose of the calculation of number of edges in the semi-CRFs model, we used $L=8$, the same as the one we used in the main paper.

We also calculated the average number of edges involved per token by averaging the average number of edges per token, over all sentences. More formally, let the number of sentences in a dataset be $N$,  the number of edges in $i$-th sentence be $E_i$, and the length of $i$-th sentence be $n_i$. Then the average number of edges involved per token $\bar{E}$ is:
\begin{equation*}
\bar{E} = \frac{\displaystyle\sum_{i=1}^{N}\frac{E_{i}}{n_{i}}}{N}.
\end{equation*}
Table \ref{tab:edges} shows the average number of edges involved per token for different models.
Since the number of edges in our models is linear in terms of $n$, the average number of edges per token will be constant, plus some small variance accounting for the boundary cases in the data.
This also explains the fact that both \textsc{dgm-s} and \textsc{dgm} models perform much faster compared to the semi-CRF model.

\begin{table}[ht!]
	\centering
	\begin{tabular}{l|ccccccc|c}
		& ABC & CNN  & MNB & NBC & P2.5 & PRI & VOA & Avg. \\ \hline
		\textsc{dgm-s} & 39.1 & 39.8 & 37.9 & 38.9 & 40.9 & 39.8 & 40.7 &39.5\\
		\textsc{dgm} & 57.3 & 62.9 & 53.7 & 56.2 & 71.3 & 59.8 & 65.5& 61.0  \\
		semi-CRFs & 117.8 & 133.9 & 110.3 & 119.8 & 172.4 & 126.4 & 144.3 & 132.1 \\
	\end{tabular}
	\caption{The average number of possible edges involved in each token when we construct the model. }
	\label{tab:edges}
\end{table}

\section{Feature Representation}
This section gives an example on the feature representation defined in Features section in the main paper. 
For illustration purpose, we take the first sentence of Figure 1 in the main paper as an example. 
The code released has the detailed feature implementation as well.

Say that we are currently at the position of word ``\textit{Ami}''. The features in the linear-chain CRF model is represented in table \ref{tab:linearf}. In semi-CRFs model, the features are defined over segment. We take the segment ``\textit{Shlomo Ben - Ami}'' as an example to describe the features in Table \ref{tab:semif}.

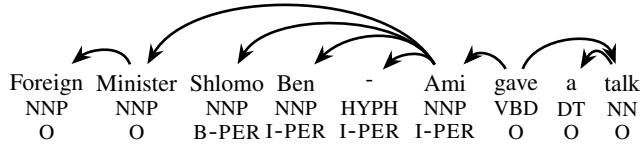
\begin{figure}[h]
	\centering
	\begin{tikzpicture}[node distance=1.0mm and 1.0mm, >=Stealth]
	\node [](anode) [] {\footnotesize Foreign};
	\node [](bnode) [right=of anode, xshift=-2mm, yshift=0.28mm] {\footnotesize Minister};
	\node [](cnode) [right=of bnode, xshift=-2mm] {\footnotesize Shlomo};
	\node [](dnode) [right=of cnode, xshift=-2mm] {\footnotesize Ben};
	\node [](enode) [right=of dnode, xshift=3mm] {\footnotesize -};
	\node [](fnode) [right=of enode, xshift=3mm] {\footnotesize Ami};
	\node [](gnode) [right=of fnode, yshift=-0.57mm] {\footnotesize gave};
	\node [](hnode) [right=of gnode, yshift=0.3mm] {\footnotesize a};
	\node [](inode) [right=of hnode, yshift=0.3mm] {\footnotesize talk};
	
	\node [](at) [below=of anode,yshift=3.0mm]{\scriptsize NNP};
	\node [](bt) [below=of bnode,yshift=2.39mm] {\scriptsize NNP};
	\node [](ct) [below=of cnode,yshift=2.39mm] {\scriptsize NNP};
	\node [](dt) [below=of dnode,yshift=2.39mm]{\scriptsize NNP};
	\node [](et) [below=of enode,yshift=1.7mm]{\scriptsize HYPH};
	\node [](ft) [below=of fnode,yshift=2.39mm]{\scriptsize NNP};
	\node [](gt) [below=of gnode,yshift=2.9mm]{\scriptsize VBD};
	\node [](ht) [below=of hnode,yshift=2.2mm]{\scriptsize DT};
	\node [](it) [below=of inode,yshift=2.29mm]{\scriptsize NN};
	
	\node [](ae) [below=of at,yshift=2.45mm]{\small \textsc{o}};
	\node [](be) [below=of bt,yshift=2.5mm] {\small \textsc{o}};
	\node [](ce) [below=of ct,yshift=2.5mm] {\small \textsc{b-per}};
	\node [](de) [below=of dt,yshift=2.5mm]{\small \textsc{i-per}};
	\node [](ee) [below=of et,yshift=2.5mm]{\small \textsc{i-per}};
	\node [](fe) [below=of ft,yshift=2.5mm]{\small \textsc{i-per}};
	\node [](ge) [below=of gt,yshift=2.5mm]{\small \textsc{o}};
	\node [](he) [below=of ht,yshift=2.5mm]{\small \textsc{o}};
	\node [](ie) [below=of it,yshift=2.5mm]{\small \textsc{o}};
	
	\draw [line width=1pt, -{Stealth[length=3.5mm, open]},->] (bnode) to [out=110,in=40, looseness=1] node [above] {} (anode);
	\draw [line width=1pt, -{Stealth[length=3.5mm, open]},->] (fnode) to [out=120,in=60, looseness=0.8] node [above] {} (bnode);
	\draw [line width=1pt, -{Stealth[length=3.5mm, open]},->] (fnode) to [out=120,in=60, looseness=0.8] node [above] {} (cnode);
	\draw [line width=1pt, -{Stealth[length=3.5mm, open]},->] (fnode) to [out=120,in=45, looseness=1] node [above] {} (dnode);
	\draw [line width=1pt, -{Stealth[length=3.5mm, open]},->] (fnode) to [out=120,in=45, looseness=1] node [above] {} (enode);
	\draw [line width=1pt, -{Stealth[length=3.5mm, open]},->] (gnode) to [out=120,in=45, looseness=1.2] node [above] {} (fnode);
	\draw [line width=1pt, -{Stealth[length=3.5mm, open]},->] (gnode) to [out=70,in=110, looseness=1] node [above] {} (inode);
	\draw [line width=1pt, -{Stealth[length=3.5mm, open]},->] (inode) to [out=120,in=60, looseness=1.8] node [above] {} (hnode);
	\end{tikzpicture} 
	\caption{The dependency tree for the sentence ``Foreign Minister Sholomo Ben-Ami gave a talk''}
	\label{fig:twoexample}
\end{figure}

\begin{table}[h!]
	\centering
	\begin{tabular}{|l|l|}
		\hline 
		Features & Examples \\ \hline
		current word & \textit{Ami} \\ \hline
		current POS & NNP \\ \hline
		previous word & \textit{-} \\ \hline
		previous POS & HYPH \\ \hline
		current word shape & Xxx \\ \hline
		previous word shape & - \\ \hline
		prefix up to length 3 & \textit{A}, \textit{Am}, \textit{Ami} \\ \hline
		suffix up to length 3 & \textit{i}, \textit{mi}, \textit{Ami} \\ \hline
		transition & \textsc{i-per} + \textsc{i-per} \\\hline 
	\end{tabular}\quad 
	\begin{tabular}{|l|l|}
		\hline 
		Dependency features & Examples \\ \hline
		current word + head & \textit{Ami}+\textit{gave} \\ \hline
		current word + head + label & \textit{Ami}+\textit{gave}+nsubj \\ \hline
		current POS + head POS & NNP+VBD \\ \hline
		current POS + head POS + label & NNP+VBD+nsubj \\ \hline
	\end{tabular}
	\caption{The features for the example sentence in the linear-chain CRFs model.}
	\label{tab:linearf}
\end{table}

\begin{table}[h!]
	\centering
	\begin{tabular}{|l|l|}
		\hline 
		Features & Examples \\ \hline
		word before segment & \textit{Minister} \\ \hline
		POS before segment & NNP \\ \hline
		word shape before segment & Xxxx \\ \hline
		word after segment & \textit{gave} \\ \hline
		POS after segment & VBD \\ \hline
		word shape after segment & xxxx \\ \hline
		prefix up to length 3 & \textit{S}, \textit{Sh}, \textit{Shl} \\ \hline
		suffix up to length 3 & \textit{i}, \textit{mi}, \textit{Ami} \\ \hline
		start word & start:+\textit{Shlomo} \\ \hline
		end word & end:+\textit{Ami} \\ \hline
		start POS & start POS:+NNP \\ \hline
		end POS & end POS:+NNP \\ \hline
		segment length & 4 \\ \hline
		transition & \textsc{o} + \textsc{per} \\\hline 
		indexed word & 1:\textit{Shlomo}, 2:\textit{Ben}, 3:\textit{-}, 4:\textit{Ami} \\ \hline
		indexed POS & 1:NNP, 2:NNP, 3:HYPH, 4:NNP \\ \hline
		indexed shape & 1:Xxxx, 2:Xxx, 3:-, 4:Xxx \\ \hline
		the whole segment & \textit{Shlomo Ben - Ami} \\ \hline
		dependency & same as dependency features in Table \ref{tab:linearf} \\ \hline
	
	\end{tabular}\quad 
%
	\caption{The features for the example sentence in the semi-CRFs model.}
	\label{tab:semif}
\end{table}

\section{Full Results with Precision, Recall and F-score}
This section presents the full results with precision, recall and F-score of all models in the paper. 
Table \ref{tab:nerresult} and Table \ref{tab:prednerresult} show the results with dependency features while Table \ref{tab:nerresultnodep} and Table \ref{tab:prednerresultnodep} show the results without dependency features. 
Recall that our \textsc{dgm-s} and \textsc{dgm} models use the dependency structure information to build the models even if we don't use the dependency features. 

\begin{table*}[h!]
	\centering
	\scalebox{0.9}{
		\begin{tabular}{l|lll|lll|lll|lll}
			& \multicolumn{3}{c|}{Linear-chain CRFs}                                   & \multicolumn{3}{c|}{Semi-Markov CRFs}                                   & \multicolumn{3}{c|}{\textsc{dgm-s}}                                              & \multicolumn{3}{c}{\textsc{dgm}}                                              \\
			& \multicolumn{1}{c}{\em P} & \multicolumn{1}{c}{\em R} & \multicolumn{1}{c|}{\em F} & \multicolumn{1}{c}{\em P} & \multicolumn{1}{c}{\em R} & \multicolumn{1}{c|}{\em F} & \multicolumn{1}{c}{\em P} & \multicolumn{1}{c}{\em R} & \multicolumn{1}{c|}{\em F} & \multicolumn{1}{c}{\em P} & \multicolumn{1}{c}{\em R} & \multicolumn{1}{c}{\em F} \\ \hline
			ABC& 	71.5&	68.9&	70.2&	71.7&	72.2&	\textbf{71.9}&	71.3&	71.5&	71.4&	72.2&	72.4&	\textbf{72.3}\\
			CNN&	76.7&	75.1&	75.9&	78.3&	78.2&	\textbf{78.2}&	77.2&	76.9&	77.0&	78.7&	78.6&	\textbf{78.6}\\
			MNB&	80.8&	71.2&	75.7&	77.4&	72.2&	\textbf{74.7}&	76.5&	70.5&	73.4&	78.8&	73.9&	\textbf{76.3}\\
			NBC&	69.0&	63.0&	65.9&	70.7&	68.2&\textbf{69.4}&	70.3&	66.7&	68.4&	70.3&	69.1&	\textbf{69.7}\\
			P2.5&	73.2&	68.6&	70.8&	75.0&	72.0&	73.5&	74.7&	70.9&	72.8&	76.7&	74.4&	\textbf{75.5}\\
			PRI&	83.9&	82.6&	83.2&	84.7&	85.5&	\textbf{85.1}&	84.8&	85.4&	\textbf{85.1}&	85.1&	85.9&	\textbf{85.5}\\
			VOA&	85.7&	83.5&	84.6&	85.6&	85.2&	85.4&	85.2&	85.1&	85.2&	87.1&	86.4&	\textbf{86.8}\\
			Overall&79.2&	76.5&	77.8&	79.9&	79.3&	79.6&	79.5&	78.6&	79.0&	80.8&	80.2&	\textbf{80.5}               
		\end{tabular}
		}
	
	\caption{Named Entity Recognition Results on the Broadcast News corpus of OntoNotes 5.0 dataset. All the models in this table are using the gold dependency information. Both \textsc{dgm-s} and \textsc{dgm} models apply the dependency information in two ways: building the model and as well as using them as features.}
	\label{tab:nerresult}
\end{table*}

\begin{table*}[h!]
	\centering
	\scalebox{0.9}{
	\begin{tabular}{l|lll|lll|lll|lll}
		& \multicolumn{3}{c|}{Linear-chain CRFs}                                   & \multicolumn{3}{c|}{Semi-Markov CRFs}                                   & \multicolumn{3}{c|}{\textsc{dgm-s}}                                              & \multicolumn{3}{c}{\textsc{dgm}}                                              \\
		& \multicolumn{1}{c}{\em P} & \multicolumn{1}{c}{\em R} & \multicolumn{1}{c|}{\em F} & \multicolumn{1}{c}{\em P} & \multicolumn{1}{c}{\em R} & \multicolumn{1}{c|}{\em F} & \multicolumn{1}{c}{\em P} & \multicolumn{1}{c}{\em R} & \multicolumn{1}{c|}{\em F} & \multicolumn{1}{c}{\em P} & \multicolumn{1}{c}{\em R} & \multicolumn{1}{c}{\em F} \\ \hline
		ABC&	70.1&	66.7&	68.4&	71.4&	71.7&	\textbf{71.6}&	70.6&	70.6&	70.6&	71.8&	72.0&	\textbf{71.9}\\
		CNN&	76.3&	74.5&	75.4&	78.0&	78.1&	\textbf{78.0}&	76.7&	76.1&	76.4&	77.6&	77.6&	\textbf{77.6}\\
		MNB&	78.6&	70.6&	74.4&	76.2&	71.0&	73.5&	76.5&	70.5&	73.4&	77.7&	73.3&	\textbf{75.4}\\
		NBC&	69.6&	63.3&	66.3&	72.5&	70.6&	\textbf{71.5}&	70.2&	67.3&	68.7&	72.0&	70.9&	\textbf{71.4}\\
		P2.5&	73.4&	68.3&	70.8&	75.2&	72.3&	\textbf{73.7}&	73.2&	69.6&	71.3&	74.7&	73.2&	\textbf{73.9}\\
		PRI&	83.9&	82.7&	83.3&	84.2&	85.0&	\textbf{84.6}&	83.7&	84.0&	\textbf{83.9}&	83.7&	84.7&	\textbf{84.2}\\
		VOA&	84.8&	82.7&	83.7&	85.4&	85.2&	\textbf{85.3}&	84.7&	84.0&	84.4&	85.3&	84.8&	\textbf{85.1}\\
		Overall&78.8&	75.9&	77.3&	79.8&	79.3&	\textbf{79.5}&	78.8&	77.6&	78.2&	79.6&	79.2&	\textbf{79.4}               
	\end{tabular}
	}
	
	\caption{Named Entity Recognition Results on the Broadcast News corpus of OntoNotes 5.0 dataset. All the models in this table are using the predicted dependency information from MaltParser.}
	\label{tab:prednerresult}
\end{table*}

\begin{table*}[t!]
	\centering
	\scalebox{0.9}{
		\begin{tabular}{l|lll|lll|lll|lll}
			& \multicolumn{3}{c|}{Linear-chain CRFs}                                   & \multicolumn{3}{c|}{Semi-Markov CRFs}                                   & \multicolumn{3}{c|}{\textsc{dgm-s}}                                              & \multicolumn{3}{c}{\textsc{dgm}}                                              \\
			& \multicolumn{1}{c}{\em P} & \multicolumn{1}{c}{\em R} & \multicolumn{1}{c|}{\em F} & \multicolumn{1}{c}{\em P} & \multicolumn{1}{c}{\em R} & \multicolumn{1}{c|}{\em F} & \multicolumn{1}{c}{\em P} & \multicolumn{1}{c}{\em R} & \multicolumn{1}{c|}{\em F} & \multicolumn{1}{c}{\em P} & \multicolumn{1}{c}{\em R} & \multicolumn{1}{c}{\em F} \\ \hline
			ABC&	67.8&	65.3&	66.5&	72.2&	72.4&	\textbf{72.3}&	69.8&	69.0&	69.4&	72.5&	72.9&	\textbf{72.7}\\
			CNN&	75.0&	73.3&	74.1&	76.7&	76.4&	76.6&	76.5&	75.7&	76.1&	77.4&	77.0&	\textbf{77.2}\\
			MNB&	77.6&	72.3&	74.9&	76.8&	73.3&	\textbf{75.0}&	76.5&	70.5&	73.4&	77.8&	73.9&	\textbf{75.8}\\
			NBC&	67.3&	63.6&	65.4&	69.8&	68.8&	\textbf{69.3}&	70.1&	66.1&	68.0&	68.5&	68.5&	\textbf{68.5}\\
			P2.5&	73.4&	68.3&	70.8&	75.2&	72.3&	73.7&	76.4&	69.0&	72.5&	77.8&	75.7&	\textbf{76.8}\\
			PRI&	83.6&	82.2&	82.9&	83.9&	84.3&	84.1&	85.0&	85.4&	85.2&	85.9&	86.5&	\textbf{86.2}\\
			VOA&	83.2&	81.4&	82.3&	83.5&	83.0&	83.3&	85.5&	84.7&	85.1&	86.1&	84.9&	\textbf{85.5}\\
			Overall&77.5&	75.1&	76.3&	78.8&	78.1&	78.5&	79.5&	77.6&	78.6&	80.3&	79.6&	\textbf{79.9}             
		\end{tabular}
		}
	
	\caption{NER results of all models without dependency features. Note that \textsc{dgm-s} and \textsc{dgm} are using the gold dependency structures in their models.}
	\label{tab:nerresultnodep}
\end{table*}

\begin{table*}[t!]
	\centering
	\scalebox{0.9}{
		\begin{tabular}{l|lll|lll|lll|lll}
			& \multicolumn{3}{c|}{Linear-chain CRFs}                                   & \multicolumn{3}{c|}{Semi-Markov CRFs}                                   & \multicolumn{3}{c|}{\textsc{dgm-s}}                                              & \multicolumn{3}{c}{\textsc{dgm}}                                              \\
			& \multicolumn{1}{c}{\em P} & \multicolumn{1}{c}{\em R} & \multicolumn{1}{c|}{\em F} & \multicolumn{1}{c}{\em P} & \multicolumn{1}{c}{\em R} & \multicolumn{1}{c|}{\em F} & \multicolumn{1}{c}{\em P} & \multicolumn{1}{c}{\em R} & \multicolumn{1}{c|}{\em F} & \multicolumn{1}{c}{\em P} & \multicolumn{1}{c}{\em R} & \multicolumn{1}{c}{\em F} \\ \hline
			ABC&	67.8&	65.3&	66.5&	72.2&	72.4&	\textbf{72.3}&	69.4&	68.8&	69.1&	71.2&	71.5&	71.3\\
			CNN&	75.0&	73.3&	74.1&	76.7&	76.4&	\textbf{76.6}&	76.1&	75.2&	75.6&	76.4&	76.0&	\textbf{76.2}\\
			MNB&	77.6&	72.3&	74.9&	76.8&	73.3&	\textbf{75.0}&	77.5&	70.5&	73.8&	78.7&	73.3&	\textbf{75.9}\\
			NBC&	67.3&	63.6&	65.4&	69.8&	68.8&	\textbf{69.3}& 69.3&	65.2&	67.2&	68.8&	68.8&	\textbf{68.8}\\
			P2.5&	73.4&	68.3&	70.8&	75.2&	72.3&	\textbf{73.7}&	75.7&	68.7&	72.0&	75.3&	73.8&	\textbf{74.6}\\
			PRI&	83.6&	82.2&	82.9&	83.9&	84.3&	84.1&	84.3&	84.7&	84.5&	84.8&	85.4&	\textbf{85.1}\\
			VOA&	83.2&	81.4&	82.3&	83.5&	83.0&	83.3&	84.7&	83.7& \textbf{84.2}&	84.9&	83.7&	\textbf{84.3}\\
			Overall&77.5&	75.1&	76.3&	78.8&	78.1&	\textbf{78.5}&	78.9&	77.0&	78.0&	79.1&	78.5&	\textbf{78.8}
		\end{tabular}
	}
	
	\caption{NER Results of all models without dependency features. Note that \textsc{dgm-s} and \textsc{dgm} are using the predicted dependency structures in their models.}
	\label{tab:prednerresultnodep}
\end{table*}

\bibliography{supp}
\bibliographystyle{plain}